\pdfoutput=1

\documentclass[11pt]{article}

\usepackage[]{EACL2023}

\usepackage{times}
\usepackage{latexsym}
\usepackage{graphicx}

\usepackage[T1]{fontenc}

\usepackage[utf8]{inputenc}

\usepackage{microtype}

\usepackage{microtype}
\usepackage{multicol}
\usepackage{soul}
\usepackage{multirow}
\usepackage{makecell}
\usepackage[utf8]{inputenc}

\usepackage{microtype}
\usepackage{graphicx}
\usepackage{soul}
\usepackage{float}
\usepackage{comment}
\usepackage{adjustbox}
\usepackage{booktabs}
\usepackage{amsmath}
\usepackage{mathabx}
\usepackage[shortlabels]{enumitem}
\usepackage[normalem]{ulem}
\usepackage{mathtools}
\usepackage{xspace}
\usepackage{array}
\usepackage{framed}
\usepackage{subcaption}
\usepackage{xcolor}

\usepackage{multicol}
\usepackage[ruled,vlined]{algorithm2e}
\usepackage{caption}
\usepackage{subcaption}
\usepackage{color,soul}
\usepackage{microtype}
\usepackage{dblfloatfix}
\usepackage{xcolor}

\usepackage{booktabs,arydshln}

\definecolor{purple}{rgb}{0.5,0,1}
\definecolor{dcyan}{rgb}{0.2,0.6,0.5}
\definecolor{darkgreen}{rgb}{0,200,0}
\definecolor{light-gray}{gray}{0.95} 
\definecolor{darkgreen}{RGB}{0,140,0}
\definecolor{darkred}{RGB}{200,0,0}
\definecolor{lightgreen}{RGB}{231,255,219}
\definecolor{lightred}{RGB}{252,231,234}
\definecolor{lightyellow}{RGB}{250,253,191}
\definecolor{DarkRed}{RGB}{130,25,0}
\usepackage{inconsolata}

\makeatletter
\def\adl@drawiv#1#2#3{%
        \hskip.5\tabcolsep
        \xleaders#3{#2.5\@tempdimb #1{1}#2.5\@tempdimb}%
                #2\z@ plus1fil minus1fil\relax
        \hskip.5\tabcolsep}
\newcommand{\cdashlinelr}[1]{%
  \noalign{\vskip\aboverulesep
           \global\let\@dashdrawstore\adl@draw
           \global\let\adl@draw\adl@drawiv}
  \cdashline{#1}
  \noalign{\global\let\adl@draw\@dashdrawstore
           \vskip\belowrulesep}}
\makeatother

\newcommand{\name}{\textsc{Natural Instructions}}
\newcommand{\namee}{\textsc{Super-NaturalInstructions}}
\newcommand*\samethanks[1][\value{footnote}]{\footnotemark[#1]}

\title{How Many Data Samples is an Additional Instruction Worth?}

\author{
Ravsehaj Singh Puri\thanks{~~Equal Contribution}$\,$ \quad Swaroop Mishra\samethanks$\,$    \quad     \textbf{Mihir Parmar} $\;$ \quad  \textbf{Chitta Baral} $\;$ 
\\
 Arizona State University, Tempe, USA
 \\
 \small{\texttt{\{rpuri8, srmishr1, mparmar3, chitta\}@asu.edu}}
}

\begin{document}
\maketitle
\begin{abstract}
Recently introduced \textit{instruction-paradigm} empowers non-expert users to leverage NLP resources by defining a new task in natural language. Instruction-tuned models have significantly outperformed multitask learning models (without instruction); however they are far from state-of-the-art task-specific models. Conventional approaches to improve model performance via creating datasets with large number of task instances or architectural changes in the model may not be feasible for non-expert users. However, they can write alternate instructions to represent an instruction task. \textit{Is Instruction-augmentation helpful?} We augment a subset of tasks in the expanded version of \name{} with additional instructions and find that it significantly improves model performance (up to 35\%), especially in the low-data regime. Our results indicate that an additional instruction can be equivalent to $\sim$200 data samples on average across tasks.\footnote{Code and dataset is available at \url{https://github.com/Ravsehajsinghpuri/Multi-Variant-Instructions}}
\end{abstract}

\section{Introduction}
Large-scale benchmarks such as Imagenet \cite{russakovsky2015imagenet}, SQuAD \cite{rajpurkar2018know} and architectural development in models such as CNNs \cite{amari2003handbook} and transformers \cite{vaswani2017attention} have propelled our progress in deep learning. However, creating high-quality benchmarks by controlling its artifacts \cite{gururangan2018annotation,Mishra2020DQIMD}, developing new models, and training them is hard for non-expert users. Recently introduced \textit{instruction-paradigm} empowers non-expert users, practitioners, and domain experts in other
fields to leverage NLP resources \cite{weller2020learning} as they now can describe their tasks in natural language without requiring to create task-specific datasets or developing models\footnote{Related work is presented in App.~\ref{app:related_work}}. Even though the instruction paradigm has led to the development of models that significantly outperform multitasking baselines, model performance has remained far behind the supervised learning model trained with task-specific data \cite{efrat2020turking, mishra2021cross}. 

Non-expert users can write multiple instructions per task each of which covers multiple perspectives spanning over a variety of linguistic features; many of these can be created automatically by replacing certain words with their synonyms without changing the overall semantics of instruction.
Can the relatively inexpensive process of instruction augmentation improve the model's performance in the \textit{instruction-paradigm}, similar to the role data-augmentation has played conventionally in machine learning \cite{feng2021survey}? \textit{Instruction-paradigm} is pivotal where it is expensive or infeasible to gather training data. How effective is instruction augmentation in low-data regimes?

Multi-variant instructions (original + augmented instructions) also can help evaluate the robustness of instruction-following models to respond to variant instructions. This is similar to the model robustness evaluation \cite{jia2019certified} that is done by creating variant data instances. Multi-variant instruction-based setup will also help gauge the true potential of instruction-following systems since in a real-world setting, users can write task instructions in many different ways.

The expanded version of \name{}~\cite{mishra2021cross,wang-etal-2022-super}\footnote{\url{https://github.com/allenai/natural-instructions}} provides a rich collection of the diverse category of tasks that covers a variety of reasoning skills, domains, and languages. This constantly evolving benchmark is growing in size with respect to time. We take 426 tasks\footnote{These were the accepted tasks in the expanded version of \name{} in September 2021. The expanded dataset is also known as \name{} v2 or \namee{}.} and creates variant instructions for each task. In \name{}, the number of instances was limited to 6500 to reduce massive data imbalance, we leverage the remaining instances of source datasets in constructing instances of our variant instruction tasks. We experiment with 3 types of learning scenarios (i) task-specific (TS), (ii) multi-task (MT), and (iii) cross-task (CT) and observe that instruction augmented models outperform their single-instruction counterparts by 17\%, 11\%, and 11\%, respectively when averaged over all experiments across the evaluation tasks. Interestingly, instruction augmentation is more effective on the low-data regime (average across 1\%, 5\%, and 10\% data) as we see a performance gain of 26\%, 16\%, and 11\% in TS, MT, and CT settings, respectively. We also quantify the contribution of each of the additional instructions and find that an additional instruction can be equivalent to $\sim$200 data samples on average across tasks.

\section{Multi-Variant Instruction Dataset}

We construct a Multi-Variant Instruction dataset on top of various tasks in \name{}. In total, our dataset has 426 different NLP tasks; each of which contains multi-variant instructions. 


\subsection{Variant Instruction Task}
\begin{figure}[ht]
    \centering
    \includegraphics[width=\columnwidth]{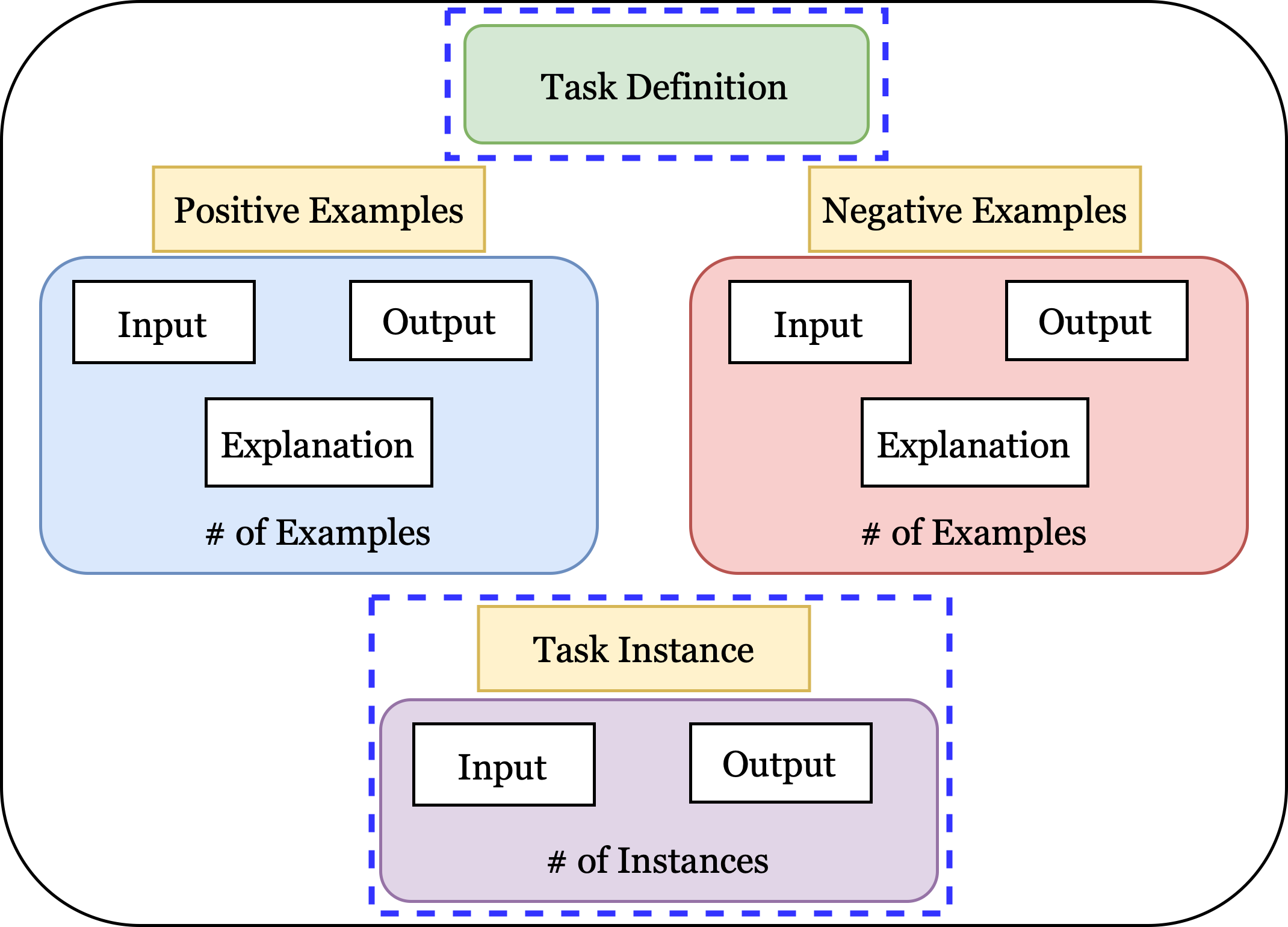}
    \caption{Schematic representation of instructional-prompts \cite{wang-etal-2022-super} - Dotted blue box represents entities that are changed in constructing variant instruction task.}
    \label{figure_instruction_schema}
\end{figure}

An instruction task in \name{} contains the definition of the task, positive examples, negative examples, and instances. Figure \ref{figure_instruction_schema} shows the schematic representation of variant instruction tasks where the blue boxes show the parts that differentiate variant instruction tasks from their original counterparts in \name{}. While constructing a variant instruction task, we alter the definition and instances of the instruction task.

\begin{table}[ht]
\large
\centering
\renewcommand{\arraystretch}{}
{
\resizebox{0.9\linewidth}{!}{
\begin{tabular}{@{}cc@{}}
\toprule
Parameter & Value \\ 
\midrule
Avg. \# of variants per task & 4.59 \\
Avg. \# of instances per task & 9510.64 \\
Avg. \# of positive examples per task & 3.15 \\
Avg. \# of negative examples per task & 2.30 \\
\bottomrule
\end{tabular}
}
}
\caption{Multi-Variant Instructions dataset statistics}
\label{tab:dataset_statistics}
\end{table}

\subsection{Dataset Creation Process}
Computer Science graduate students who participated in the data creation process are asked to create as many variant instruction tasks as possible. They are instructed to change the definition (without changing the semantic meaning of the definition in the original task) and instances (by random sampling from the set of instances in the source dataset which is not part of instruction tasks in \name{}. They are allowed to use automated tools such as Semantic Control \cite{ross2021tailor}, Text Style Transfer \cite{reif2021recipe}, NL-Augmenter \cite{dhole2021nl}. Sometimes, the participants create variant instruction tasks manually. Table \ref{app:example_instruction_1} and Table \ref{app:example_instruction_2} in App. \ref{app:example_variants} illustrates examples of alternate definitions across variant instructions created for our dataset.

\subsection{Dataset Properties and Statistics}
Table \ref{tab:dataset_statistics} shows the statistics of our meta-dataset. Note that, variant instruction tasks contain all instances from \name{}, so the average number of instances per task is higher than 6500 (which is a constraint in \name). We describe various attributes of our dataset in the following.

\subsubsection{Semantic Textual Similarity}
\begin{figure}[ht]
    \centering
    \includegraphics[width=\columnwidth]{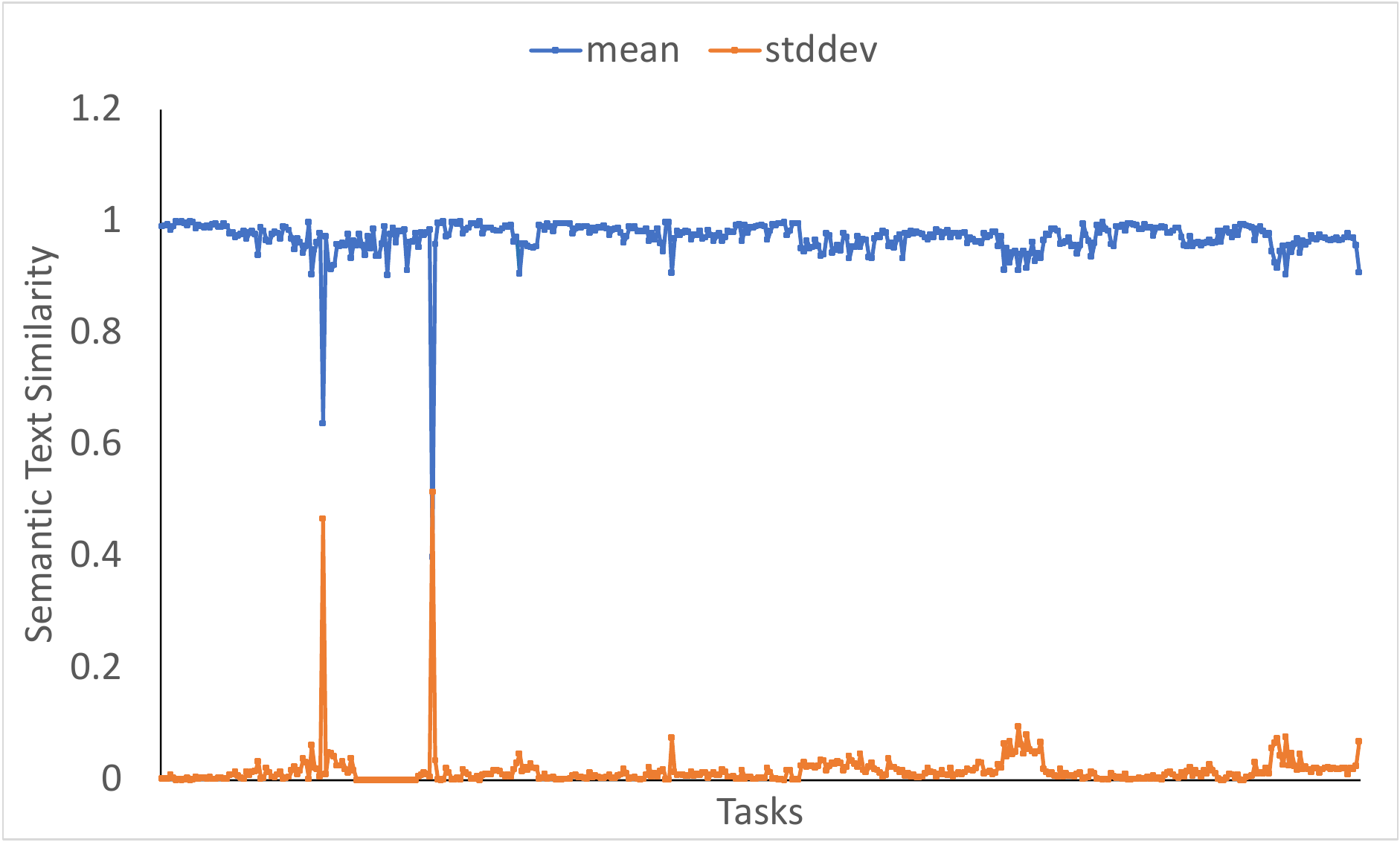}
    \caption{Semantic text similarity between original instruction and its variants.}
    \label{fig:figure_semantic_text_similarity_chart}
\end{figure}

Semantic Textual Similarity (STS) should be high between original instruction and augmented instructions as they represent the same task. We compute the pair-wise STS score between definitions of original instruction and variant instructions. Figure \ref{fig:figure_semantic_text_similarity_chart} shows the mean and SD of STS score between original instruction and its variants across 426 tasks. More detail is presented in App. \ref{app:dataset_additional_details}.

\begin{table*}[t]
\large
\centering
\resizebox{0.95\textwidth}{!}{
\begin{tabular}{@{}cccc@{}}
\toprule

\textbf{Task ID} &  \textbf{Task Name} & \textbf{Task Category} & \textbf{\# of Variants} \\ 
\midrule
 task010 & winogrande\_answer\_generation & Answer Generation & 8\\
 task011 & winogrande\_question\_modification\_object & Text Modification & 8\\
 task012 & winogrande\_question\_modification\_person & Text Modification & 8\\
 task017 & qasc\_question\_generation & Question Generation & 8\\
 task018 & qasc\_answer\_generation & Answer Generation & 8\\
 task020 & essential\_terms\_answering\_incomplete\_questions & Classification & 8\\
 task028 & multirc\_correct\_answer\_single\_sentence & Answer Generation & 3\\
 task058 & babi\_t1\_single\_supporting\_fact\_answer\_generation & Answer Generation & 5\\
\bottomrule
\end{tabular}
}
\caption{Number of variant instructions for 8 different tasks}
\label{tab:table_multi_task_descriptions}
\end{table*}

\paragraph{Analysis of dataset properties} From all dataset properties, we can observe that STS score is higher for almost all the tasks. This indicates that all augmented variants are semantically similar to the original instruction. Moreover, we can see a significant variation in terms of word dissimilarity and length of definitions (see App. \ref{app:dataset_additional_details}). From this, we can conclude that the variants created in our meta-dataset for each task have sufficient variations in terms of words and length yet sustain semantic similarity with original instruction.

\section{Experimental Setup}

\subsection{Models}
BART-base \cite{lewis2019bart} and T5-base \cite{raffel2020exploring} models are used with default hyper parameters from Huggingface \cite{wolf2019huggingface} to perform experiments.
We use Single Instruction (SI) learning as baseline where only original instruction is used to fine-tune the model. We propose Multi-Variant Instruction (MVI) learning where variants are used to fine-tune models. We use the same number of instances for both original and variant instruction learning to accurately gauge the importance of additional instructions.

\subsection{Experiments}

We perform three experiments: (1) Task-Specific, (2) Multi-Task, and (3) Cross-Task. All experiments are performed using 1\%, 5\%, 10\%, 50\% and 100\% instances from the task for fine-tuning. Here, we divide instances into train, test and dev splits by randomly sampling in the ratio 70\%, 20\% and 10\%, respectively. Evaluation is performed on the test set of original instructions. As SI is dependent on \name{} which has exactly one instruction per task, this limits our experiments to use only one instruction in the SI setting while comparing it with MVI which has multiple variant instructions.

\paragraph{Task-Specific} Here, we fine-tune the baseline and our model on one task and evaluate on the same task. We have performed task-specific learning on 3 different tasks - winogrande\_answer\_generation, winogrande\_question\_modification\_person, and qasc\_answer\_generation. In addition, we also analyze two different tasks in other task categories like tweetqa\_question\_generation and odd-man-out\_classification\_no\_category for generation and classification tasks respectively.

\paragraph{Multi-Task} To perform multi-task learning, we use 8 different tasks spanning across 4 different categories. Table \ref{tab:table_multi_task_descriptions} shows the different number of variant instructions for 8 tasks and their categories. In this setting, we fine-tune the baseline and our model on all 8 tasks combined and evaluate on each task. However, we use only two positive and two negative examples to satisfy the maximum token limit of the BART-base.

\paragraph{Cross-Task} Here, we fine-tune the model on a set of tasks and evaluate on a different set of tasks. Here, we use 274 different tasks for training by sampling 10\% instances from each task and evaluate on a set of 8 tasks which are the same as in the multi-task setup. In addition to sampling instances, we also sampled number of tasks by taking 1\%, 5\%, 10\%, 50\%, and 100\% tasks. We also investigate the extent of cross-task generalization in low-data regimes; we do this by randomly sampling 1\%, 5\%, and 10\% instances for fine-tuning.

\paragraph{Metric} We use the Rouge-L metric \cite{lin2004rouge} for evaluation in all our experiments, following the evaluation in \name{}.

\section{Results and Analysis}

\subsection{Experimental Results}

\begin{figure}[ht]
    \centering
    \includegraphics[width=\columnwidth]{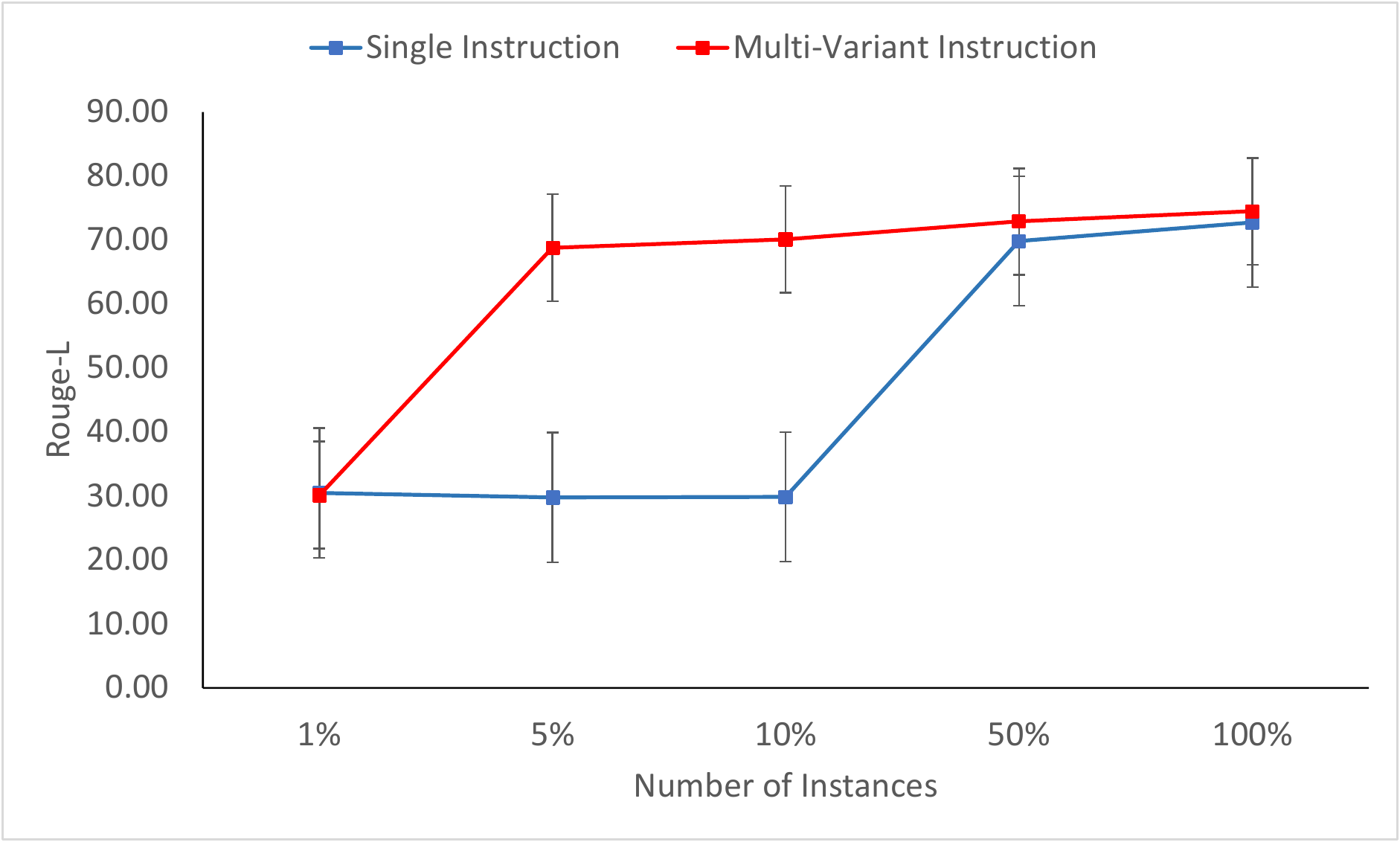}
    \caption{Comparison across SI and MVI learning in task-specific setting; Results are averaged over 3 tasks.}
    \label{fig:single_task_results}
\end{figure}

\paragraph{Task-Specific} Figure \ref{fig:single_task_results} shows the comparison between SI and MVI across a different number of instances sampled for fine-tuning. From this, we can observe that MVI outperforms SI by 17\% on average. The performance difference between MVI and SI increases to 26\% in a low data regime (average performance with 1\%, 5\%, and 10\% instances for fine-tuning). We observe similar results for the additional 2 tasks we have analyzed (present in App. \ref{app:single_task}).

\begin{figure}[ht]
    \centering
    \includegraphics[width=\columnwidth]{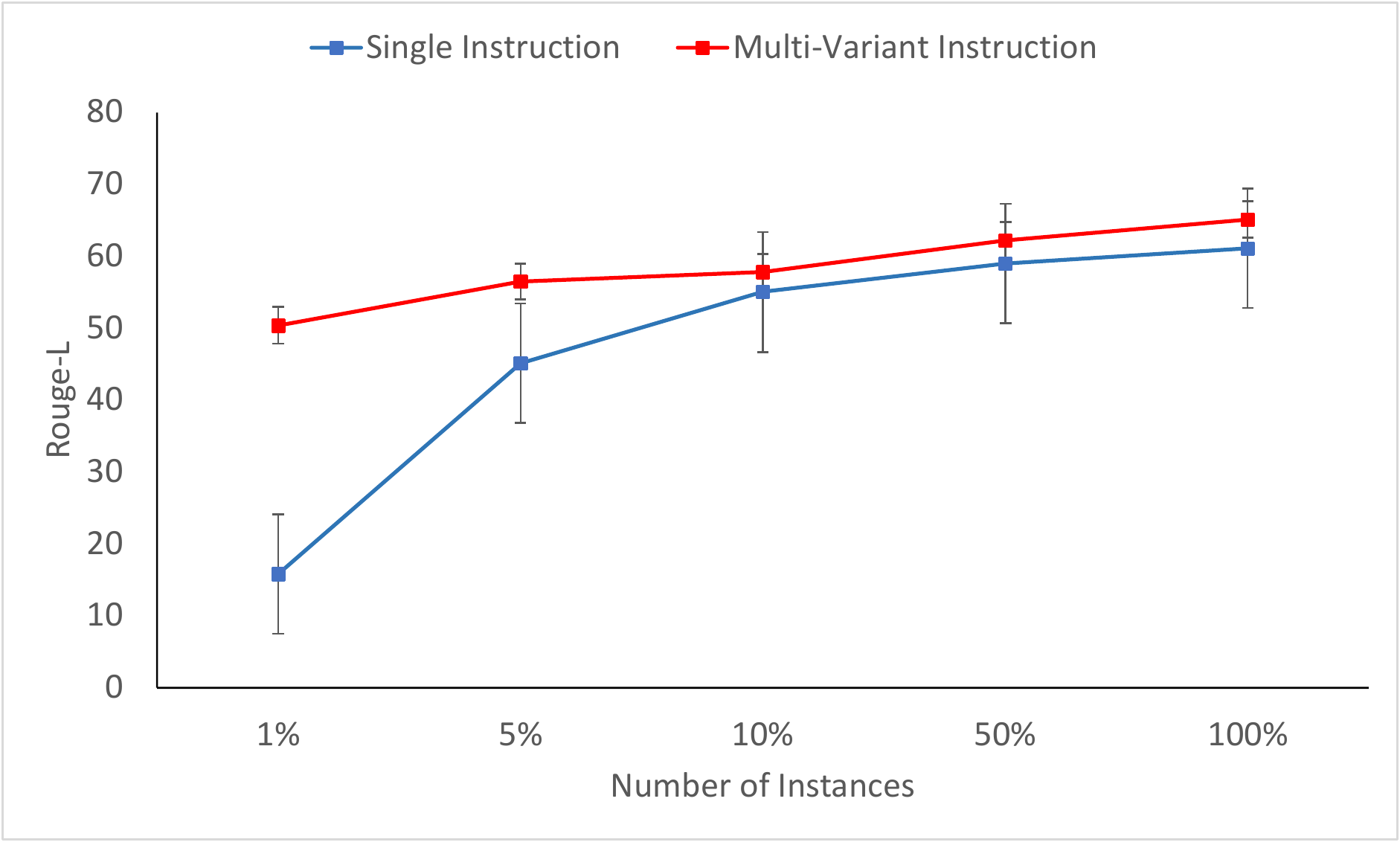}
    \caption{Comparison across SI and MVI learning in multi-task setting by varying number of instances.}
    \label{fig:figure_multi_task_line_chart}
\end{figure}

\paragraph{Multi-Task} Figure \ref{fig:figure_multi_task_line_chart} presents the comparison between SI and MVI for multi-task setting. We can observe that MVI outperforms SI by 11\% on an average. Moreover, we can see higher improvement in low data regime ($\sim16\%$). Our model achieves high performance boost ($\sim$35\%) at 1\% instances setting. App. \ref{app:multi_task} contains more details.

\begin{figure}[ht]
    \centering
    \includegraphics[width=\columnwidth]{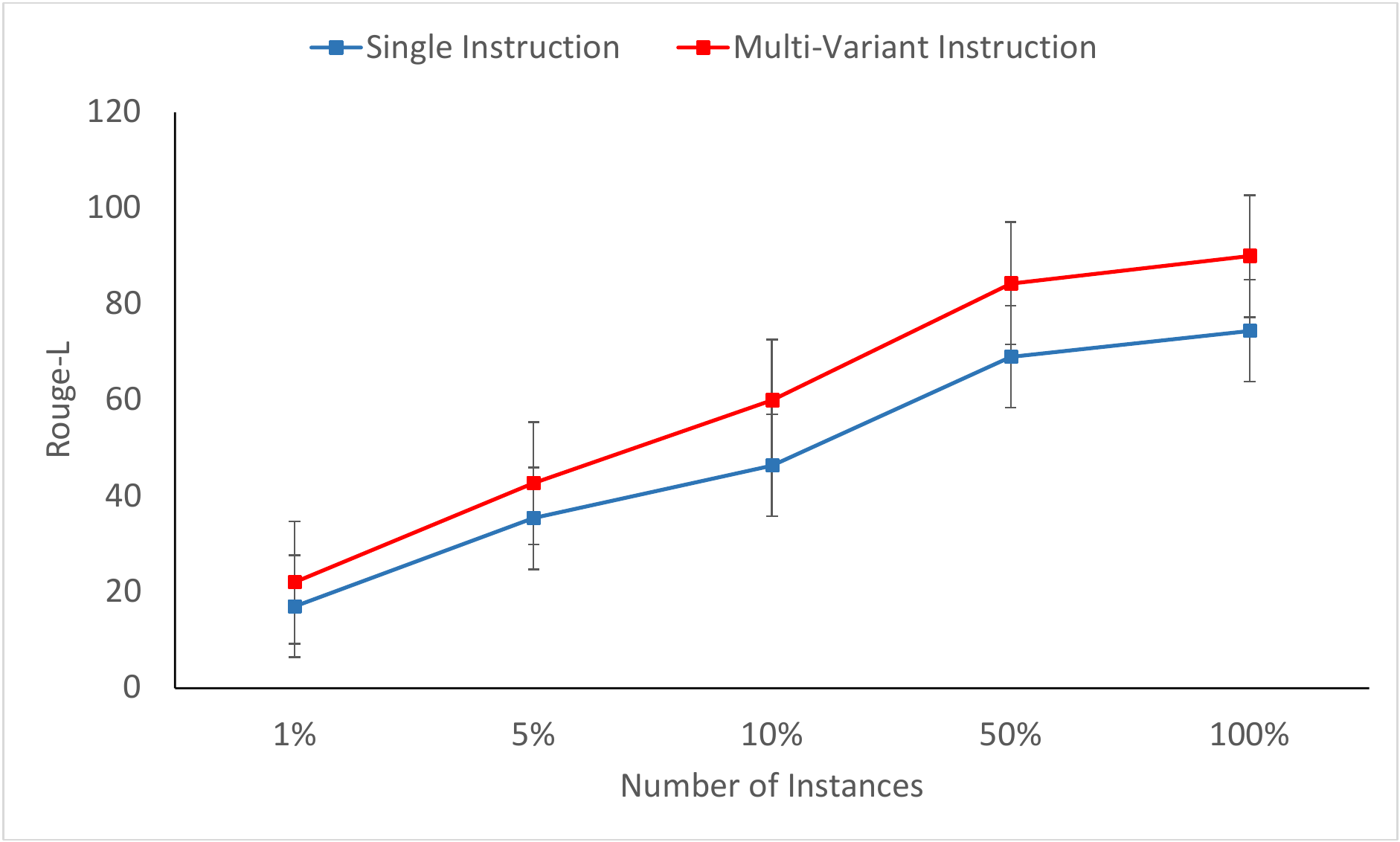}
    \caption{Comparison between SI and MVI learning in cross-task setting by varying number of instances and fixing number of tasks to 100\%.}
    \label{fig:figure_cross_task_line_chart_100}
\end{figure}

\paragraph{Cross-Task} 
Figure \ref{fig:figure_cross_task_line_chart_100} shows a comparison between SI and MVI for 100\% tasks in cross-task setting (see Figure \ref{app:cross_task_line_chart} in App. \ref{app:cross_task} for other settings). We can observe that MVI outperforms SI by 9\% on an average. App. \ref{app:cross_task} contains more details.

\subsection{Analysis}

\paragraph{How Many Data Samples is a Variant Instruction Worth?} We calculate the contribution of an additional instruction with respect to data samples in the following way: we calculate model performance for BART-base in MVI with 5\% instances. We interpolate the model performance plot in SI to find out the percentage of instances needed to match performance in MVI (with 5\% instances). We divide the average number of instance difference by average number of instruction variants to get the number that indicates worth of an additional instruction in terms of data samples. Using the above described procedure, we calculate the contribution for additional instruction in all three settings and summarize the results in Table \ref{tab:table_instruction_equivalence}. We use MVI performance with 5\% instances as the base because a typical instruction-paradigm is designed in a "low-data regime" where non-expert users can teach a task to a model without requiring to create a dataset. However, we also calculated the instruction-equivalence using MVI with 10\% instances as the base and report the results in Table \ref{tab:table_instruction_equivalence}. On an average across TS, MT and CT, we conclude that an additional variant instruction alone is worth $\sim$200 instances.

\begin{table}[ht]
\large
\centering
\renewcommand{\arraystretch}{}
{
\resizebox{0.9\linewidth}{!}{
\begin{tabular}{@{}ccccc@{}}
\toprule
Base & Task-Specific & Multi-Task & Cross-Task & Average \\ 
\midrule
5\%  & 456.2 & 94.1 & 152.3 & 234.2 \\
10\% & 460.4 & 58.2 & 279.6 & 266.1 \\
\bottomrule
\end{tabular}
}
}
\caption{Weight of each additional instruction in terms of number of data samples across task-specific, multi-task and cross-task settings.}
\label{tab:table_instruction_equivalence}
\end{table}

\paragraph{Equal Data Analysis} We believe that each instruction variant is equivalent to $\sim$200 data instances. To show this by experiment, we perform equal data analysis and observe that model trained using our approach shows competitive performance compared to single-instruction learning by using only N/V instances where N is the total number of instances in the original task and V is the number of instruction variants for this task. See App. \ref{app:equal_data_analysis} for more details.

\paragraph{Is Model Robust to Instruction Perturbations?} Here, we introduce 3 perturbations while testing SI and MVI: (1) we perturb the instruction by removing the task definition, (2) we perturb the instruction by changing the order of positive and negative examples by placing positive examples followed by negative different from training setup, and (3) we perturb the instruction by removing all positive and negative examples from the test set. We evaluate the model's robustness across these perturbations (performance change while the change in instruction) which are excluded from the training data. Here, Table \ref{tab:table_perturbation_results_single_task} for task-specific setting on T5-base (see Table \ref{app:table_perturbation_results_multi_task} in App. \ref{app:robustness_analysis} for multi-task results). We can clearly observe that our approach is robust to all three instruction perturbations whereas model trained with single-instruction learning is not able to perform equally well on perturbed test sets compared to its original test counterpart. A similar trend is observed in the multi-task setting as well (see App. \ref{app:robustness_analysis}).

\begin{table}[t]
    \centering 
    \setlength\tabcolsep{4.0pt}
    \footnotesize
\renewcommand{\arraystretch}{1.2}
{
\resizebox{\linewidth}{!}{
\begin{tabular}{@{}ccccccccc@{}}
\toprule

\multirow{2}{*}{\# of Instances} &  \multicolumn{2}{c}{SI} & \multicolumn{2}{c}{Perturbation 1} & \multicolumn{2}{c}{Perturbation 2} & \multicolumn{2}{c}{Perturbation 3}\\ 
\cmidrule(lr){2-9} 
~& Original & Ours & Original & Ours & Original & Ours & Original & Ours\\ 
\bottomrule
 1\% & 0.90	& 25.21 & 1.60 & 18.03 & 1.02 & 23.16 & 5.12 & 9.71 \\
 5\% & 0.98	& 75.72 & 2.18 & 75.32 & 1.36 & 75.50 & 5.52 & 74.26 \\
 10\% & 50.88 & 78.20 & 20.76 & 78.07 & 50.49 & 78.37 & 40.31 & 77.22 \\
 50\% & 76.55 & 82.16 & 68.88 & 82.15 & 76.50 & 82.16 & 75.34 & 81.92 \\
 100\% & 79.38 & 83.16 & 73.51 & 82.97 & 79.34 & 83.12 & 78.71 & 82.40 \\
\bottomrule
\end{tabular}
}
}
\caption{Comparison of performance in task-specific setting across SI and MVI learning.}
\label{tab:table_perturbation_results_single_task}
\end{table}
\vspace{-0.08cm}
\section{Conclusion}

We introduced instruction augmentation to improve existing LMs in terms of improving performance and usability to non-expert users. To this extent, we created multi-variant instructions for 426 NLP tasks. Our experiment results show that instruction augmentation improves model performance in task-specific, multi-task and cross-task learning paradigms. We find that instruction augmentation is more effective in low-data regime. Our results further indicate that an additional instruction can be equivalent to $\sim$200 instances on an average. We hope our work will bring more attention to developing unconventional techniques (beyond dataset creation and model training) to empower non-expert users to leverage NLP resources and teach a task without having domain knowledge.

\section*{Limitations}
We use BART-base and T5-base for all our experiments, however, we wish to experiment with different language models in future to show the benefit of our approach. Our analysis includes only tasks in English language, hence, it is important to see if our approach can be extended to non-English tasks as well. We feel that developing diverse instruction augmentation techniques will be pivotal to achieving more improvements as future research.

\bibliography{anthology,custom}
\bibliographystyle{acl_natbib}

\clearpage

\appendix

\section{Related Work}
\label{app:related_work}
\paragraph{Prompt Learning} Due to the success of large LMs, research paradigm in ML/DL has been shifted to prompt-based learning to achieve generalization and eliminate the need of creating task-specific models and large scale datasets \cite{liu2021pre}. Past attempts have been made using prompt-based learning to solve various tasks including text classification \cite{yin2019benchmarking}, Natural Language Inference (NLI) \cite{schick2020exploiting}, Question Answering (QA) \cite{jiang2020can}, Information Extraction (IE) \cite{chen2021adaprompt, cui2021template} and many more \cite{liu2021pre}. Recently, T0 model \cite{sanh2021multitask} is proposed which uses prompts to achieve zero-shot generalization across various NLP tasks. We were motivated by the work of \citet{le2021many} which shows that prompting is often worth 100s of data points on average. Our work instead focuses on instructions that are often different in terms of length, language, and capacity to represent a task~\cite{wang-etal-2022-super}. Additionally, in contrast to prior works, we focus on the use of automatic methods for instruction augmentation and evaluate its efficacy across low-data to high-data regime in task-specific, multi-task, cross-task setups. 

\paragraph{Instruction Learning} \citet{efrat2020turking} studies whether existing LMs understands instructions. After that, many works have been proposed to show that models follow language instructions \cite{hase2021can, ye2021zero, gupta2021towards, zhong2021adapting}. Furthermore, \cite{weller2020learning} has developed a framework that focuses on developing NLP systems that solve new tasks after reading their descriptions. \citet{mishra2021cross} has proposed natural language instructions for cross-task generalization of LMs. Along with that, PromptSource and FLAN \cite{wei2021finetuned, sanh2021multitask} were built for leveraging instructions and achieving zero-shot generalization on unseen tasks. Moreover, \citet{parmar-etal-2022-boxbart} shows the effectiveness of instructions in multi-task settings for the biomedical domain. \citet{mishra2021reframing} discuss the impact of task instruction reframing on model response. \citet{min2021metaicl} introduce a framework to better understand in-context learning. \citet{ouyang2022training} propose the InstructGPT model that is fine-tuned with human feedback to follow instructions. \citet{wang2022instructionner} has developed instruction-based multi-task framework for few-shot Named Entity Recognition (NER) tasks. In addition, many approaches have been proposed to improve model performance using instructions \cite{wu2022ai, lin2021few, wang-etal-2022-super, luo2022biotabqa, kuznia-etal-2022-less, patel-etal-2022-question, mishra2022help}.

\section{Example of Variants}
\label{app:example_variants}

Table \ref{app:example_instruction_1} and Table \ref{app:example_instruction_2} show the examples of different variants created from the task117\_afs\_argument\_similarity\_gun\_control and task018\_qasc\_answer\_generation respectively.

\begin{table*}
    \small
    \centering
    \begin{tabular}{p{0.42cm}p{14cm}}
    \toprule
         & \textit{Original instruction along with its augmented variant instructions} \\ 
        \midrule
        \parbox[t]{2mm}{\multirow{3}{*}{\rotatebox[origin=c]{90}{
         \textsc{\textcolor{magenta}{\parbox{3.7cm}{\small \centering Original \\ Instruction}}}
         }}}
        & \textcolor{blue}{\textit{\textbf{Definition:}}} 
        \textit{We would like you to classify each of the following sets of argument pairs (discussing Gun Control)  into either SIMILAR or NOT SIMILAR. A pair of arguments is considered SIMILAR if the arguments are about the same FACET (making the same argument), and is considered NOT SIMILAR if they do not have the same FACET. A FACET is a low level issue that often reoccurs in many arguments in support of the author's stance or in attacking the other author's position.} \\
        \rule{0pt}{0.4cm} & \textcolor{blue}{\textit{\textbf{Negative Examples:}}} \newline \textbf{Input}: <input> \quad \textbf{Output}: <output> \quad \textbf{Explanation}: <explanation> \\
        \rule{0pt}{0.4cm} & \textcolor{blue}{\textit{\textbf{Positive Examples:}}} \newline \textbf{Input}: <input> \quad \textbf{Output}: <output> \quad \textbf{Explanation}: <explanation> \\ 
        
        \midrule
        \parbox[t]{2mm}{\multirow{3}{*}{\rotatebox[origin=c]{90}{
         \textsc{\textcolor{magenta}{\parbox{3.0cm}{\small \centering Variant \\ Instruction 1}}}
         }}}
        & \textcolor{blue}{\textit{\textbf{Definition:}}} 
        \textit{Each of the following sets of argument pairs (on the topic of Gun Control) should be classified as SIMILAR or NOT SIMILAR. If the arguments are about the same FACET (making the same argument), they are deemed SIMILAR; otherwise, they are NOT SIMILAR. A FACET is a low-level problem that appears frequently in many arguments in favor of the author's position or in opposition to the position of the other author.} \\
        \rule{0pt}{0.4cm} & \textcolor{blue}{\textit{\textbf{Negative Examples:}}} \newline \textbf{Input}: <input> \quad \textbf{Output}: <output> \quad \textbf{Explanation}: <explanation> \\
        \rule{0pt}{0.4cm} & \textcolor{blue}{\textit{\textbf{Positive Examples:}}} \newline \textbf{Input}: <input> \quad \textbf{Output}: <output> \quad \textbf{Explanation}: <explanation> \\ 
        
        \midrule
        \parbox[t]{2mm}{\multirow{3}{*}{\rotatebox[origin=c]{90}{
         \textsc{\textcolor{magenta}{\parbox{3.7cm}{\small \centering Variant \\ Instruction 2}}}
         }}}
        & \textcolor{blue}{\textit{\textbf{Definition:}}} 
        \textit{Please classify the following sets of argument pairs (discussing the Gun Control) as SIMILAR or NOT SIMILAR. If the arguments are about the same FACET (making the same argument), they are regarded SIMILAR; if they are not, they are considered NOT SIMILAR. A FACET is a low-level problem that frequently recurs in numerous arguments in favor of the author's position or in opposition to the position of the other author.} \\
        \rule{0pt}{0.4cm} & \textcolor{blue}{\textit{\textbf{Negative Examples:}}} \newline \textbf{Input}: <input> \quad \textbf{Output}: <output> \quad \textbf{Explanation}: <explanation> \\
        \rule{0pt}{0.4cm} & \textcolor{blue}{\textit{\textbf{Positive Examples:}}} \newline \textbf{Input}: <input> \quad \textbf{Output}: <output> \quad \textbf{Explanation}: <explanation> \\ 

        \midrule
        \parbox[t]{2mm}{\multirow{3}{*}{\rotatebox[origin=c]{90}{
         \textsc{\textcolor{magenta}{\parbox{2.2cm}{\small \centering Variant \\ Instruction 3}}}
         }}}
        & \textcolor{blue}{\textit{\textbf{Definition:}}} 
        \textit{Two arguments are SIMILAR if they are making the same case related to author's position, else they are NOT SIMILAR. Your task is to classify any 2 arguments as SIMILAR or NOT SIMILAR.} \\
        \rule{0pt}{0.4cm} & \textcolor{blue}{\textit{\textbf{Negative Examples:}}} \newline \textbf{Input}: <input> \quad \textbf{Output}: <output> \quad \textbf{Explanation}: <explanation> \\
        \rule{0pt}{0.4cm} & \textcolor{blue}{\textit{\textbf{Positive Examples:}}} \newline \textbf{Input}: <input> \quad \textbf{Output}: <output> \quad \textbf{Explanation}: <explanation> \\  
        
        \midrule
        \parbox[t]{2mm}{\multirow{3}{*}{\rotatebox[origin=c]{90}{
         \textsc{\textcolor{magenta}{\parbox{3.0cm}{\small \centering Variant \\ Instruction 4}}}
         }}}
        & \textcolor{blue}{\textit{\textbf{Definition:}}} 
        \textit{Each of the following sets of argument pairs (discussing the Gun Control) should be classified as SIMILAR or NOT SIMILAR. If the arguments are about the same FACET (making the same argument), they are regarded SIMILAR; otherwise, they are NOT SIMILAR. A FACET is a low-level issue that appears frequently in many arguments in support of the author's position or in opposition to the position of the other author.} \\
        \rule{0pt}{0.4cm} & \textcolor{blue}{\textit{\textbf{Negative Examples:}}} \newline \textbf{Input}: <input> \quad \textbf{Output}: <output> \quad \textbf{Explanation}: <explanation> \\
        \rule{0pt}{0.4cm} & \textcolor{blue}{\textit{\textbf{Positive Examples:}}} \newline \textbf{Input}: <input> \quad \textbf{Output}: <output> \quad \textbf{Explanation}: <explanation> \\ 
        \bottomrule
\end{tabular}
\caption{Example of an instruction for a classification task with its variant instructions; these belong to the task117\_afs\_argument\_similarity\_gun\_control.}
    \label{app:example_instruction_1}
\end{table*}
\begin{table*}[]
    \small
    \centering
    \begin{tabular}{p{0.42cm}p{14cm}}
    \toprule
         & \textit{Original instruction along with its augmented variant instructions} \\ 
        \midrule
        \parbox[t]{2mm}{\multirow{3}{*}{\rotatebox[origin=c]{90}{
         \textsc{\textcolor{magenta}{\parbox{3.7cm}{\small \centering Original \\ Instruction}}}
         }}}
        & \textcolor{blue}{\textit{\textbf{Definition:}}} 
        \textit{Write a correct answer to the given question based on its associated fact. Make sure that your answer is contained in the associated fact. Things to avoid: Don't be creative and introduce any new word that is not mentioned in the associated fact! Remember that, the associated fact has been rearranged to form the question. So, the correct answer words must lie within the associated fact. Emphasis \& Caution: The correct answer can be a word, phrase, or even a sentence.} \\
        \rule{0pt}{0.4cm} & \textcolor{blue}{\textit{\textbf{Negative Examples:}}} \newline \textbf{Input}: <input> \quad \textbf{Output}: <output> \quad \textbf{Explanation}: <explanation> \\
        \rule{0pt}{0.4cm} & \textcolor{blue}{\textit{\textbf{Positive Examples:}}} \newline \textbf{Input}: <input> \quad \textbf{Output}: <output> \quad \textbf{Explanation}: <explanation> \\ 
        
        \midrule
        \parbox[t]{2mm}{\multirow{3}{*}{\rotatebox[origin=c]{90}{
         \textsc{\textcolor{magenta}{\parbox{3.7cm}{\small \centering Variant \\ Instruction 1}}}
         }}}
        & \textcolor{blue}{\textit{\textbf{Definition:}}} 
        \textit{Handwriting a rectify reply to the given issue based on its related fact. Make sure that your replying is contained in the associated fact. Aspects to avoidance: Don't be creativity and introduces any nouveau word that is not alluded in the associated doing! Recall that, the linked doing has been restructured to forma the question. Thus, the corrects replying words needs lie within the associated doing. Focuses \& Discretion: The exact replying can be a word, phrase, or even a penalties.} \\
        \rule{0pt}{0.4cm} & \textcolor{blue}{\textit{\textbf{Negative Examples:}}} \newline \textbf{Input}: <input> \quad \textbf{Output}: <output> \quad \textbf{Explanation}: <explanation> \\
        \rule{0pt}{0.4cm} & \textcolor{blue}{\textit{\textbf{Positive Examples:}}} \newline \textbf{Input}: <input> \quad \textbf{Output}: <output> \quad \textbf{Explanation}: <explanation> \\ 
        
        \midrule
        \parbox[t]{2mm}{\multirow{3}{*}{\rotatebox[origin=c]{90}{
         \textsc{\textcolor{magenta}{\parbox{3.7cm}{\small \centering Variant \\ Instruction 2}}}
         }}}
        & \textcolor{blue}{\textit{\textbf{Definition:}}} 
        \textit{Write a correcting responding to the gave question bases on its associated fact. Make persuaded that your answering is contained in the associated facto.Matters to shirk: Don't be inventive and introduce any nouveau word that is not referred in the associated fact! Recollect that, the associated fact has been redesigned to forma the issue. Therefore, the accurate responses words owes lying inside the associated doing. Concentrating \& Circumspect: The correcting responses can be a word, phrase, or even a punishments.} \\
        \rule{0pt}{0.4cm} & \textcolor{blue}{\textit{\textbf{Negative Examples:}}} \newline \textbf{Input}: <input> \quad \textbf{Output}: <output> \quad \textbf{Explanation}: <explanation> \\
        \rule{0pt}{0.4cm} & \textcolor{blue}{\textit{\textbf{Positive Examples:}}} \newline \textbf{Input}: <input> \quad \textbf{Output}: <output> \quad \textbf{Explanation}: <explanation> \\ 

        \midrule
        \parbox[t]{2mm}{\multirow{3}{*}{\rotatebox[origin=c]{90}{
         \textsc{\textcolor{magenta}{\parbox{3.7cm}{\small \centering Variant \\ Instruction 3}}}
         }}}
        & \textcolor{blue}{\textit{\textbf{Definition:}}} 
        \textit{Write a corrects answer to the afforded issue founded on its associated fact. Deliver sure that your replied is contain in the linked fact. Things to shirk: Don't be creative and introduce any novel word that is not alluded in the associated fact! Remind that, the associated doing has been redesigned to forme the question. Accordingly, the correcting reply phrases needs lied indoors the linked fact. Concentrates \& Caveat: The corrects response can be a word, phrase, or even a condemnation.} \\
        \rule{0pt}{0.4cm} & \textcolor{blue}{\textit{\textbf{Negative Examples:}}} \newline \textbf{Input}: <input> \quad \textbf{Output}: <output> \quad \textbf{Explanation}: <explanation> \\
        \rule{0pt}{0.4cm} & \textcolor{blue}{\textit{\textbf{Positive Examples:}}} \newline \textbf{Input}: <input> \quad \textbf{Output}: <output> \quad \textbf{Explanation}: <explanation> \\  
        
        \midrule
        \parbox[t]{2mm}{\multirow{3}{*}{\rotatebox[origin=c]{90}{
         \textsc{\textcolor{magenta}{\parbox{3.7cm}{\small \centering Variant \\ Instruction 4}}}
         }}}
        & \textcolor{blue}{\textit{\textbf{Definition:}}} 
        \textit{Writing a accurate responded to the yielded matter founded on its associated fact. Deliver sure that your reply is contained in the associated doing. Aspects to avoidance: Don't be creative and introduce any newer word that is not talked in the associated facto! Recall that, the associated fact has been rearranged to form the issue. Thereby, the corrects responding phrase gotta lie within the related doing. Focus \& Circumspect: The correct responding can be a word, expression, or even a sentences.} \\
        \rule{0pt}{0.4cm} & \textcolor{blue}{\textit{\textbf{Negative Examples:}}} \newline \textbf{Input}: <input> \quad \textbf{Output}: <output> \quad \textbf{Explanation}: <explanation> \\
        \rule{0pt}{0.4cm} & \textcolor{blue}{\textit{\textbf{Positive Examples:}}} \newline \textbf{Input}: <input> \quad \textbf{Output}: <output> \quad \textbf{Explanation}: <explanation> \\ 
        
        \midrule
        \parbox[t]{2mm}{\multirow{3}{*}{\rotatebox[origin=c]{90}{
         \textsc{\textcolor{magenta}{\parbox{3.7cm}{\small \centering Variant \\ Instruction 5}}}
         }}}
        & \textcolor{blue}{\textit{\textbf{Definition:}}} 
        \textit{Writing a correct answers to the granted question bases on its associated doing. Make sure that your respond is contained in the associated doing. Matters to shirk: Don't be creative and introduces any novo word that is not referenced in the associated facto! Remind that, the associated fact has been reconfigured to forms the question. So, the corrects respond words ought lies within the related doing. Concentrate \& Careful: The accurate reply can be a word, phrase, or yet a sentences.} \\
        \rule{0pt}{0.4cm} & \textcolor{blue}{\textit{\textbf{Negative Examples:}}} \newline \textbf{Input}: <input> \quad \textbf{Output}: <output> \quad \textbf{Explanation}: <explanation> \\
        \rule{0pt}{0.4cm} & \textcolor{blue}{\textit{\textbf{Positive Examples:}}} \newline \textbf{Input}: <input> \quad \textbf{Output}: <output> \quad \textbf{Explanation}: <explanation> \\ 
        \bottomrule
\end{tabular}
\caption{Example of an instruction for an answer generation task with its variant instructions - task018\_qasc\_answer\_generation}
    \label{app:example_instruction_2}
\end{table*}

\section{Multi-Variant Dataset Additional Details}
\label{app:dataset_additional_details}

\subsection{Semantic Textual Similarity}
We use en\_core\_web\_md semantic similarity model of SpaCy to compute STS in our experiments. We also calculate STS score between definitions of variants of the same task. At the end, we calculate their mean and Standard Deviation (SD) for each task. 

In the plot, the two exception points are task058 (Answer generation task based on babi dataset \cite{weston2015towards}) and task097 (Structured text generation task based on SCAN dataset \cite{lake2018generalization}) where the original instructions are very long and the variant task contains a short definition which causes the strong variation in STS. We also discuss the Word-Level Dissimilarity and Length Diversity properties of our dataset below.

\subsection{Word-Level Dissimilarity} 
\begin{figure}[ht]
    \centering
    \includegraphics[width=\columnwidth]{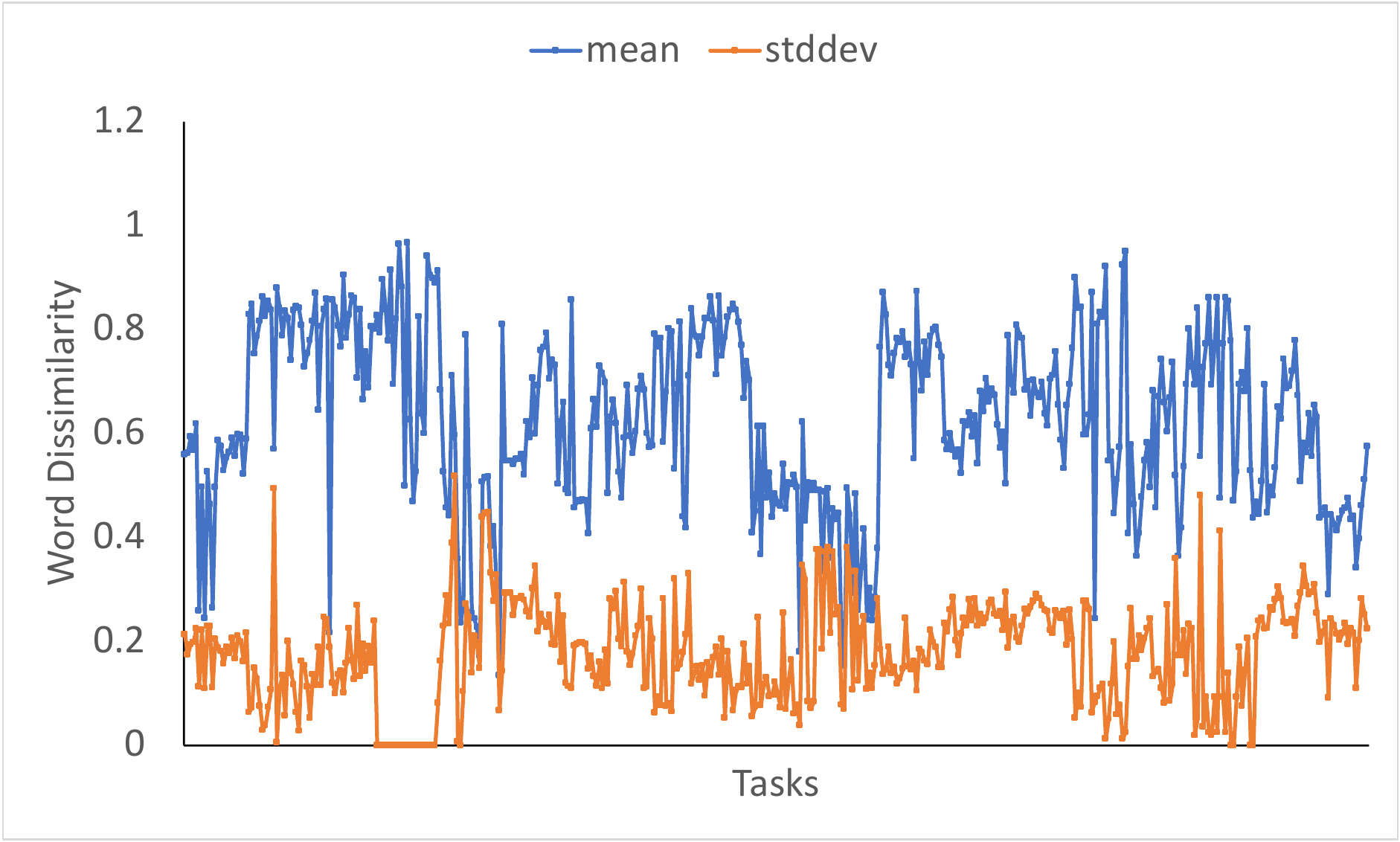}
    \caption{Word-level dissimilarity between original instruction and its variants.}
    \label{fig:figure_word_dissimilarity_chart}
\end{figure}

To show the quality and diversity of variant instructions, we calculate the pair-wise edit distance between the definition of the original instruction and its variant instructions. We also calculate distance between definitions of variant instructions of the same task, further normalize by the highest distance to obtain a dissimilarity score. We compute the mean and SD of these scores for each task and show it in Figure \ref{fig:figure_word_dissimilarity_chart}.

\subsection{Length Diversity}
\begin{figure}[ht]
    \centering
    \includegraphics[width=\columnwidth]{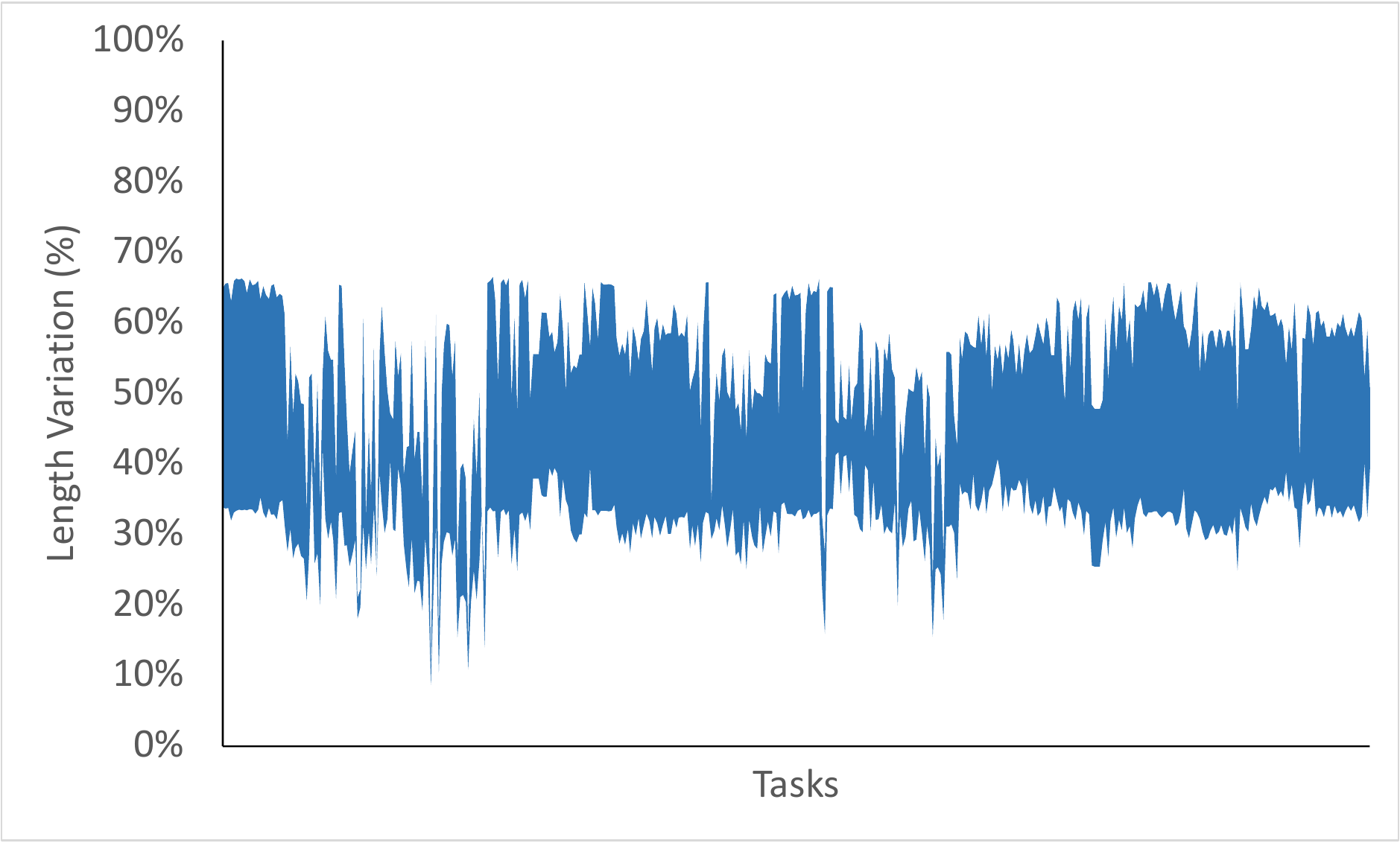}
    \caption{Definition length variation between original instruction and its variants.}
    \label{fig:figure_definition_length_variation_chart}
\end{figure}

It is necessary to see how task definition lengths vary between original instructions and their variants. To understand this, we compute the percentage difference between the length of the maximum instruction definition and the minimum instruction definition for each task and show it in Figure \ref{fig:figure_definition_length_variation_chart}.

\section{Task-Specific Results}
\label{app:single_task}

Table \ref{tab:table_single_task_combined} shows the results for task-specific experiments for task010\_winogrande\_answer\_generation, task012\_winogrande\_question\_modification\_person and task018\_qasc\_answer\_generation. We also performed experiments for other task categories like task210\_tweetqa\_question\_generation and task113\_odd-man-out\_classification\_no\_category for generation and classification tasks respectively and summarize our results in Table \ref{tab:table_single_task_additional}. From the average results, we can observe that multi-variant instruction learning helps model to improve performance in task-specific learning.

\begin{table*}[t]
    \setlength\tabcolsep{3.0pt}
    \setlength{\belowcaptionskip}{-10pt}
    \centering
    \footnotesize
\renewcommand{\arraystretch}{1.2}
{
\resizebox{0.85\linewidth}{!}{
\begin{tabular}{@{}ccccccccccc@{}}
\toprule
\multirow{3}{*}{\# of Instances} &  \multicolumn{4}{c}{BART-base} & \multicolumn{4}{c}{T5-base}\\
\cmidrule{2-9}
& \multicolumn{2}{c}{SI} & \multicolumn{2}{c}{MVI} & \multicolumn{2}{c}{SI} & \multicolumn{2}{c}{MVI} \\ 
\cmidrule{2-9}
~& Original & Ours & Original & Ours & Original & Ours & Original & Ours\\ 
\midrule
~& & & & task\_010 & & &\\
\midrule
 1\% & 0.00 & 0.00 & 0.00 & 0.02 & 0.04 & 13.71 & 0.16 & 11.26\\
 5\% & 0.00 & 36.75 & 0.06 & 37.07 & 0.01 & 46.44 & 0.14 & 44.69\\
 10\% & 0.23 & 39.17 & 0.15 & 38.26 & 12.03 & 53.03 & 9.05 & 52.60\\
 50\% & 37.00 & 43.02 & 25.40 & 42.54 & 48.11 & 64.94 & 46.01 & 64.80\\
 100\% & 41.97 & 45.65 & 33.84 & 45.50 & 55.67 & 67.49 & 53.74 & 66.92\\
\midrule
~& & & & task\_012 & & &\\
\midrule
 1\% & 84.48 & 83.54 & 75.45 & 82.66 & 0.07 & 0.00 & 6.20 & 6.17\\
 5\% & 84.73 & 90.68 & 74.52 & 90.68 & 0.05 & 90.90 & 6.17 & 90.87\\
 10\% & 84.81 & 90.61 & 75.47 & 90.60 & 79.62 & 90.99 & 62.69 & 90.99\\
 50\% & 90.29 & 90.49 & 85.65 & 90.48 & 90.92 & 90.77 & 90.81 & 90.81\\
 100\% & 90.84 & 90.50 & 88.47 & 90.52 & 91.02 & 90.75 & 90.87 & 90.80 \\
\midrule
~& & & & task\_018 & & &\\
\midrule
 1\% & 7.05 & 6.92 & 4.36 & 5.27 & 2.57 & 61.92 & 3.02 & 58.53 \\
 5\% & 4.65 & 79.07 & 3.42 & 79.55 & 2.89 & 89.84 & 3.80 & 89.99\\
 10\% & 4.72 & 80.59 & 3.68 & 80.95 & 61.00 & 90.57 & 56.28 & 90.56\\
 50\% & 82.43 & 85.23 & 81.36 & 85.20 & 90.63 & 90.76 & 90.86 & 90.79\\
 100\% & 85.58 & 87.37 & 84.90 & 87.52 & 91.44 & 91.25 & 91.41 & 91.11\\
\midrule
~& & & & Average & & &\\
\midrule
 1\% & 30.51 & 30.15 & 26.60 & 29.32 & 0.90 & 25.21 & 3.12 & 25.32 \\
 5\% & 29.79 & 68.83 & 26.00 & 69.10 & 0.98 & 75.72 & 3.37 & 75.18\\
 10\% & 29.92 & 70.12 & 26.43 & 69.94 & 50.88 & 78.20 & 42.67 & 78.05\\
 50\% & 69.91 & 72.91 & 64.14 & 72.74 & 76.55 & 82.16 & 75.89 & 82.13\\
 100\% & 72.80 & 74.51 & 69.07 & 74.51 & 79.38 & 83.16 & 78.68 & 82.94\\
\bottomrule
\end{tabular}
}
}
\caption{Comparison of performance in single-task setting across single-instruction and multi-variant instruction learning. SI: Single-Instruction, MVI: Multi-Variant Instruction.}
\label{tab:table_single_task_combined}
\end{table*}
\begin{table*}[t]
    \setlength\tabcolsep{3.0pt}
    \setlength{\belowcaptionskip}{-10pt}
    \centering
    \footnotesize
\renewcommand{\arraystretch}{1.2}
{
\resizebox{0.5\linewidth}{!}{
\begin{tabular}{ccccc}
\toprule
\# of Instances & SI & MVI & SI & MVI \\ 
\midrule
&\multicolumn{2}{c}{task\_210} & \multicolumn{2}{c}{task\_113} \\
\midrule
 1\% & 13.37 & 12.25 & 3.00 & 3.85 \\
 5\% & 13.50 & 25.92 & 4.77 & 15.26 \\
 10\% & 14.67 & 27.14 & 4.00 & 30.77 \\
 50\% & 27.88 & 41.06 & 41.72 & 81.80 \\
 100\% & 37.24 & 44.10 & 66.73 & 98.10 \\
\bottomrule
\end{tabular}
}
}
\caption{Comparison of performance in task-specific setting across single-instruction and multi-variant instruction learning. SI: Single-Instruction}
\label{tab:table_single_task_additional}
\end{table*}

\section{Multi-Task Results}
\label{app:multi_task}

The results for multi-task learning experiments are shown in Table \ref{tab:table_multi_task}.

\begin{table*}[t]
    \setlength\tabcolsep{3.0pt}
    \setlength{\belowcaptionskip}{-10pt}
    \centering
    \footnotesize
\renewcommand{\arraystretch}{1.2}
{
\resizebox{0.9\linewidth}{!}{
\begin{tabular}{@{}ccccccccccc@{}}

\toprule
\multirow{3}{*}{\# of Instances} &  \multicolumn{4}{c}{BART-base} & \multicolumn{4}{c}{T5-base}\\
\cmidrule{2-9}
& \multicolumn{2}{c}{SI} & \multicolumn{2}{c}{MVI} & \multicolumn{2}{c}{SI} & \multicolumn{2}{c}{MVI} \\ 
\cmidrule{2-9}
~& Original & Ours & Original & Ours & Original & Ours & Original & Ours\\ 
\midrule
 1\% & 15.84 & 50.40 & 14.97 & 51.88 & 7.34	& 34.53 & 6.11 & 33.61\\
 5\% & 45.13 & 56.49 & 44.24 & 57.71 & 32.01 & 62.61 & 19.88 & 62.87\\
 10\% & 55.03 & 57.80 & 51.67 & 58.70 & 46.93 & 63.61 & 39.76 & 63.98\\
 50\% & 59.01 & 62.21 & 57.37 & 62.06 & 63.38 & 66.16 & 57.11 & 66.76\\
 100\% & 61.08 & 65.13 & 58.58 & 65.09 & 64.99 & 67.15 & 59.35 & 67.38 \\
\bottomrule
\end{tabular}
}
}
\caption{Comparison of performance in multi-task setting across single-instruction and multi-variant instruction learning. SI: Single-Instruction, MVI: Multi-Variant Instruction}
\label{tab:table_multi_task}
\end{table*}

\section{Cross-Task Results}
\label{app:cross_task}

The results for cross-task learning experiments are shown in Table \ref{tab:table_cross_task}. Figure \ref{app:cross_task_line_chart} compares single-instruction learning and our approach in cross-task setting.

\section{Equal Data Analysis}
\label{app:equal_data_analysis}

We keep the original number of instances in SI learning, however, reduce the number of instances used in MVI learning by sampling N/V number of instances randomly for each task where N is the total number of instances in the original task and V is the number of instruction variants for this task. We perform these experiments in both task-specific and multi-task settings using BART-base. Table \ref{tab:table_equal_data_analysis_results} summarizes the results of these experiments, and we can observe that the model trained using our approach shows competitive performance compared to single-instruction learning by using only N/V instances.
\begin{table*}[ht]
    \centering 
    \setlength\tabcolsep{4.0pt}
    \footnotesize
\renewcommand{\arraystretch}{1.2}
{
\resizebox{0.6\textwidth}{!}{
\begin{tabular}{@{}cccccc@{}}
\toprule

\multirow{2}{*}{\# of Instances} &  \multicolumn{2}{c}{Single Task} & \multicolumn{2}{c}{Multi Task} & \\ 
\cmidrule(lr){2-5} 
~& Original & Ours & Original & Ours\\ 
\midrule
 1\% & 10.81 & 7.32 & 6.35 & 0.82\\
 5\% & 20.86 & 19.42 & 4.21 & 6.31\\
 10\% & 57.22 & 51.36 & 59.95 & 49.42\\
 50\% & 76.53 & 72.75 & 84.54 & 79.74\\
 100\% & 78.36 & 60.15 & 86.55 & 82.02\\
 \midrule
 Average & 48.76 & 42.20 & 48.32 & 43.66\\
\bottomrule
\end{tabular}
}
}
\caption{Comparison of performance in task-specific (average across 3 tasks) and multi-task settings.}
\label{tab:table_equal_data_analysis_results}
\end{table*}

\begin{figure*}
  \begin{subfigure}[b]{0.5\linewidth}
    \centering
    \includegraphics[width=0.9\linewidth]{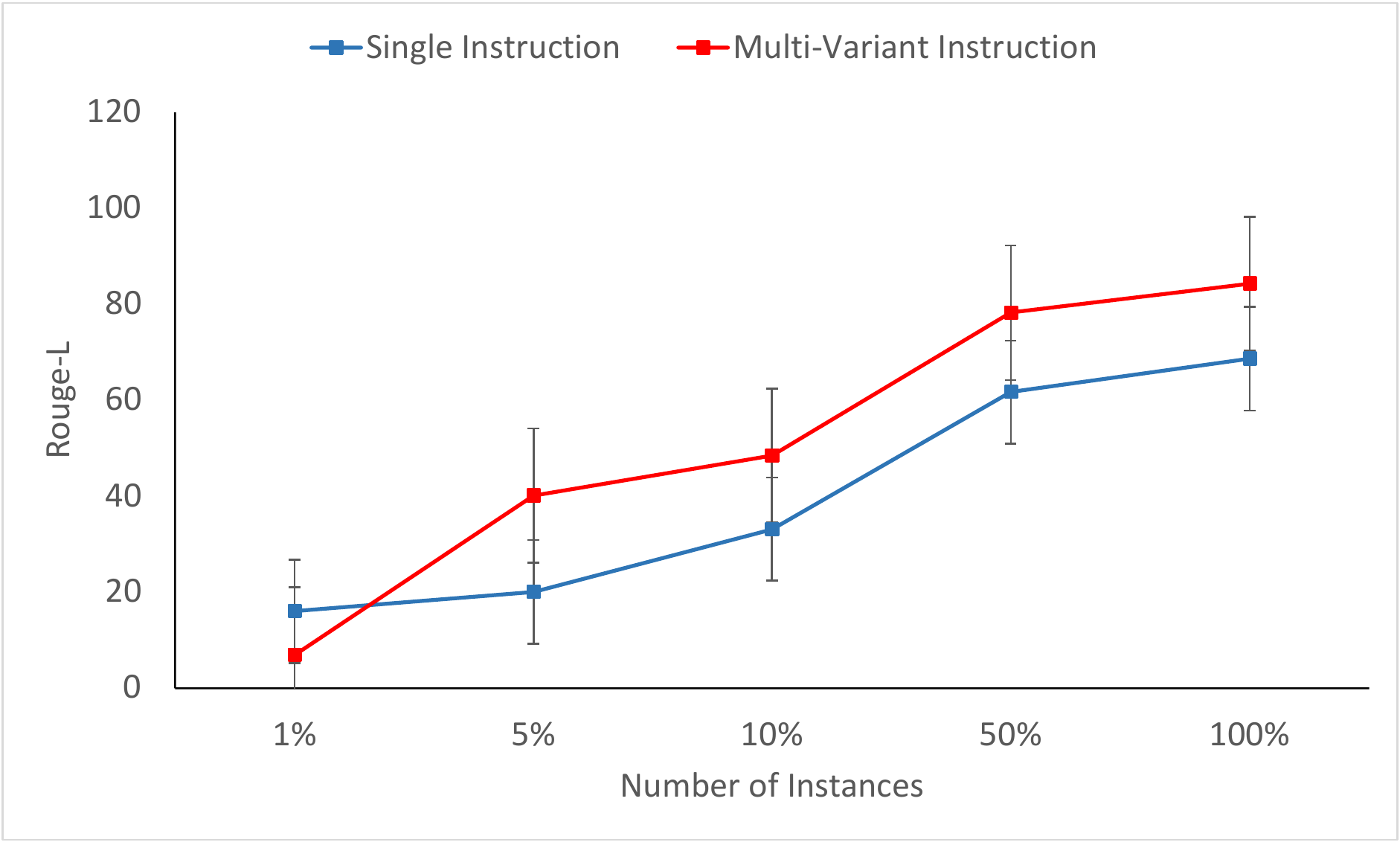} 
    \caption{fixing number of tasks to 1\%} 
    \label{cross_task_line_chart:a} 
    \vspace{4ex}
  \end{subfigure}
  \begin{subfigure}[b]{0.5\linewidth}
    \centering
    \includegraphics[width=0.9\linewidth]{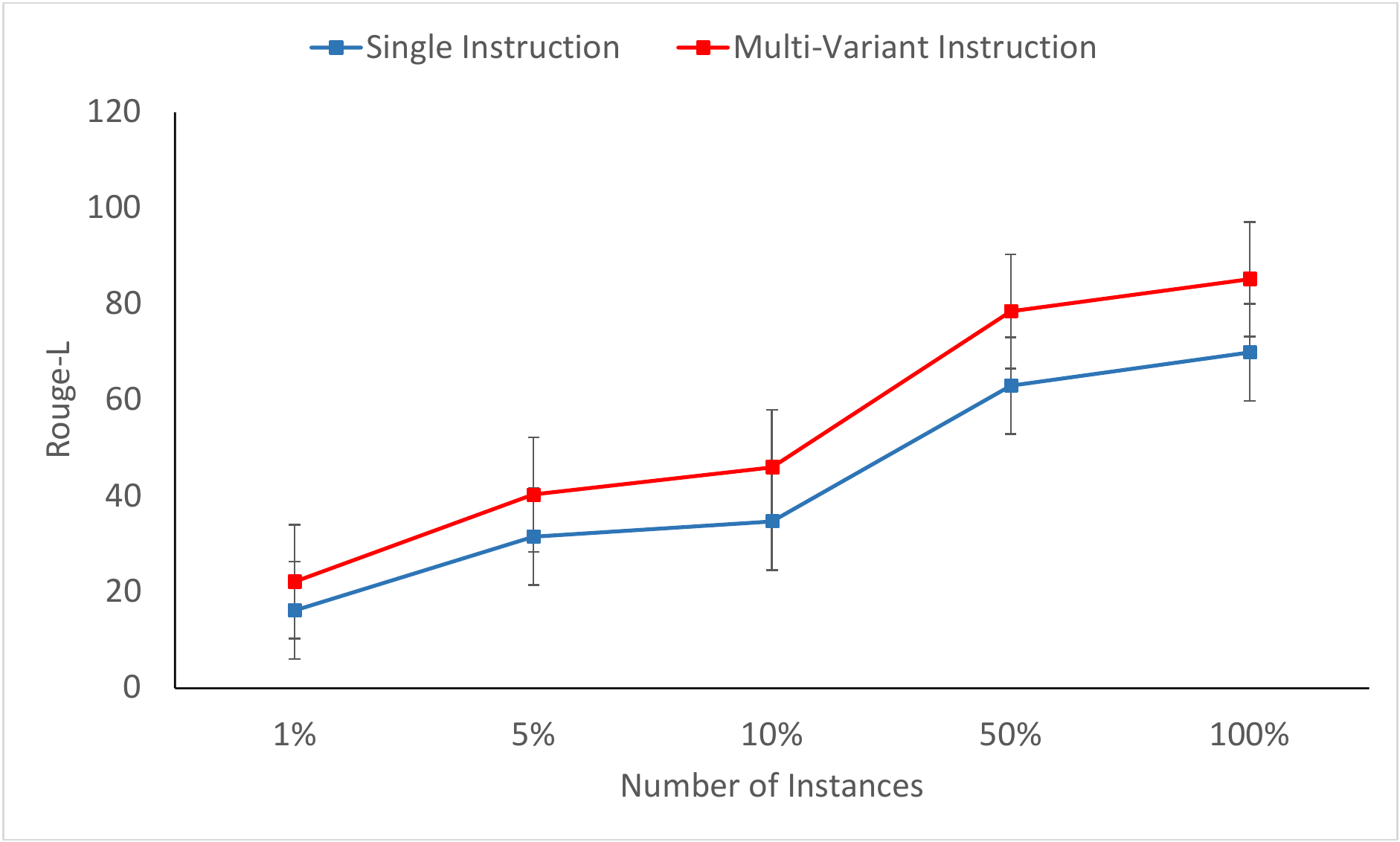} 
    \caption{fixing number of tasks to 5\%} 
    \label{cross_task_line_chart:b} 
    \vspace{4ex}
  \end{subfigure} 
  \begin{subfigure}[b]{0.5\linewidth}
    \centering
    \includegraphics[width=0.9\linewidth]{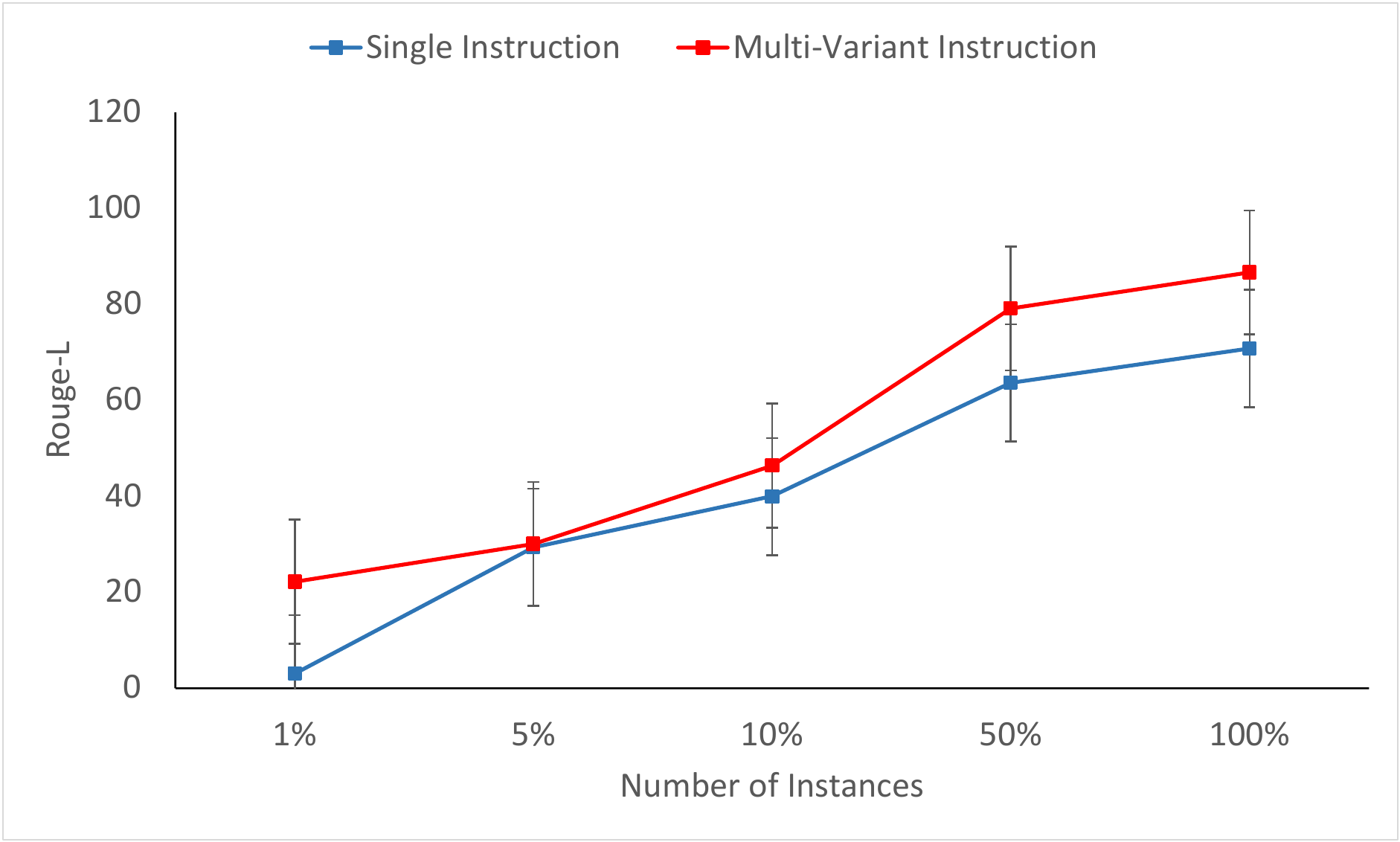} 
    \caption{fixing number of tasks to 10\%} 
    \label{cross_task_line_chart:c} 
    \vspace{4ex}
  \end{subfigure}
  \begin{subfigure}[b]{0.5\linewidth}
    \centering
    \includegraphics[width=0.9\linewidth]{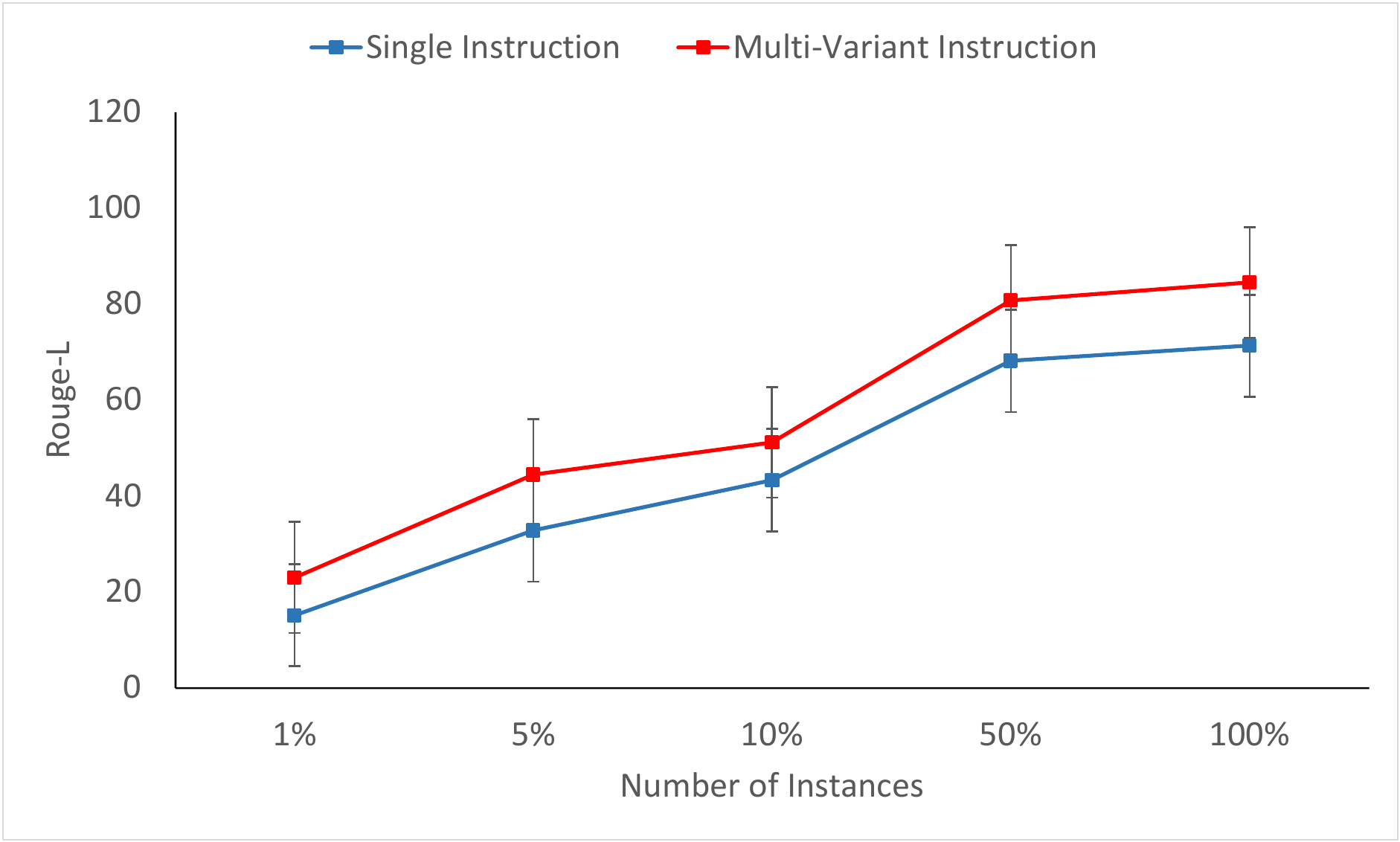} 
    \caption{fixing number of tasks to 50\%}
    \label{cross_task_line_chart:d} 
    \vspace{4ex}
  \end{subfigure}
  \caption{Comparison of performance across SI and MVI learning in cross-task setting by varying number of instances and tasks. Evaluation is performed on the test set of original instructions.}
  \label{app:cross_task_line_chart} 
\end{figure*}

The results for cross-task learning experiments are shown in Table \ref{tab:table_cross_task}. Figure \ref{app:cross_task_line_chart} compares single-instruction learning and our approach in cross-task setting.

\section{Robustness Analysis}
\label{app:robustness_analysis}

\begin{figure*}[ht] 
  \begin{subfigure}[b]{0.5\linewidth}
    \centering
    \includegraphics[width=0.9\linewidth]{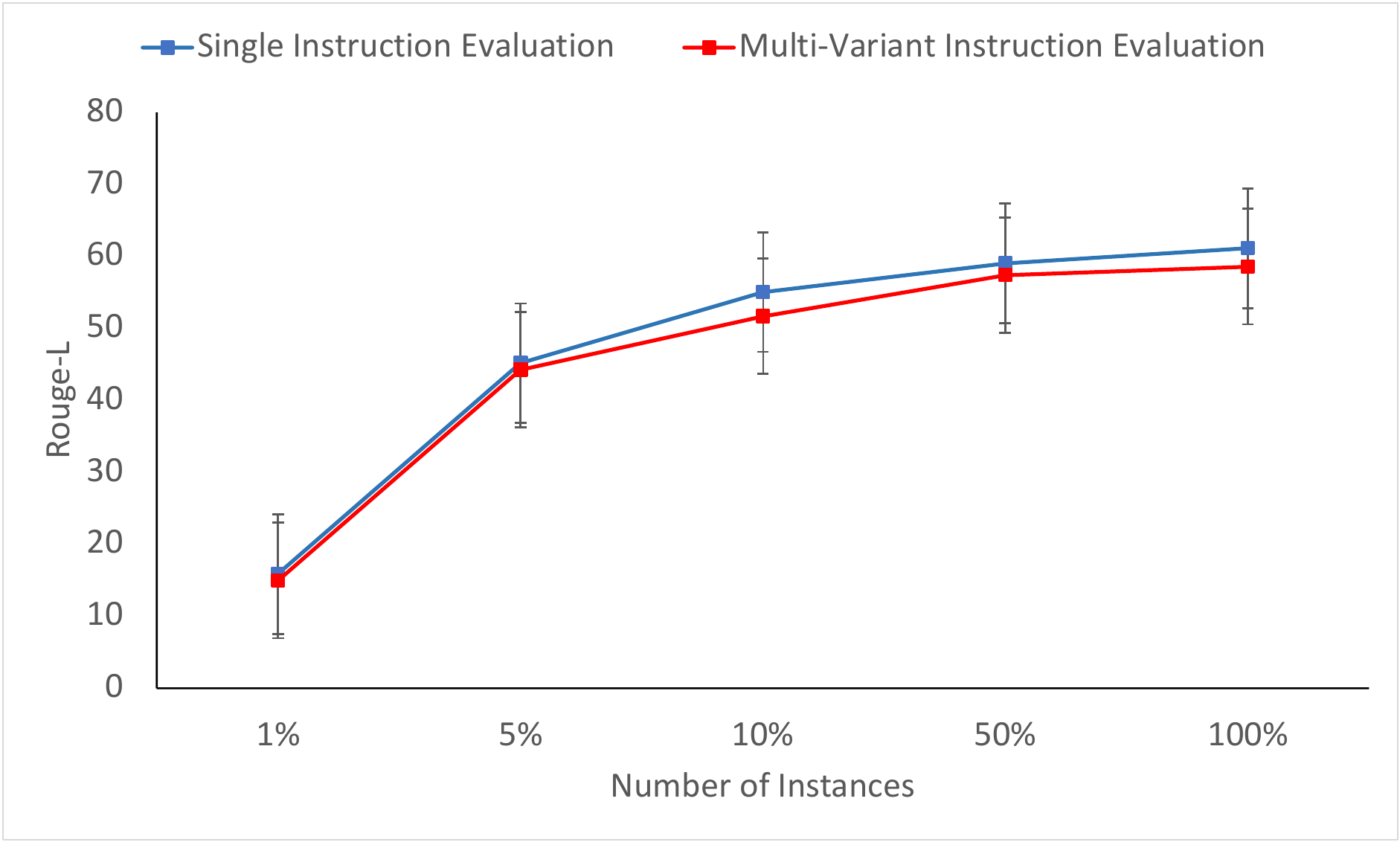} 
    \caption{Multi-task SI learning} 
    \label{robustness_line_chart:a} 
    \vspace{4ex}
  \end{subfigure}
  \begin{subfigure}[b]{0.5\linewidth}
    \centering
    \includegraphics[width=0.9\linewidth]{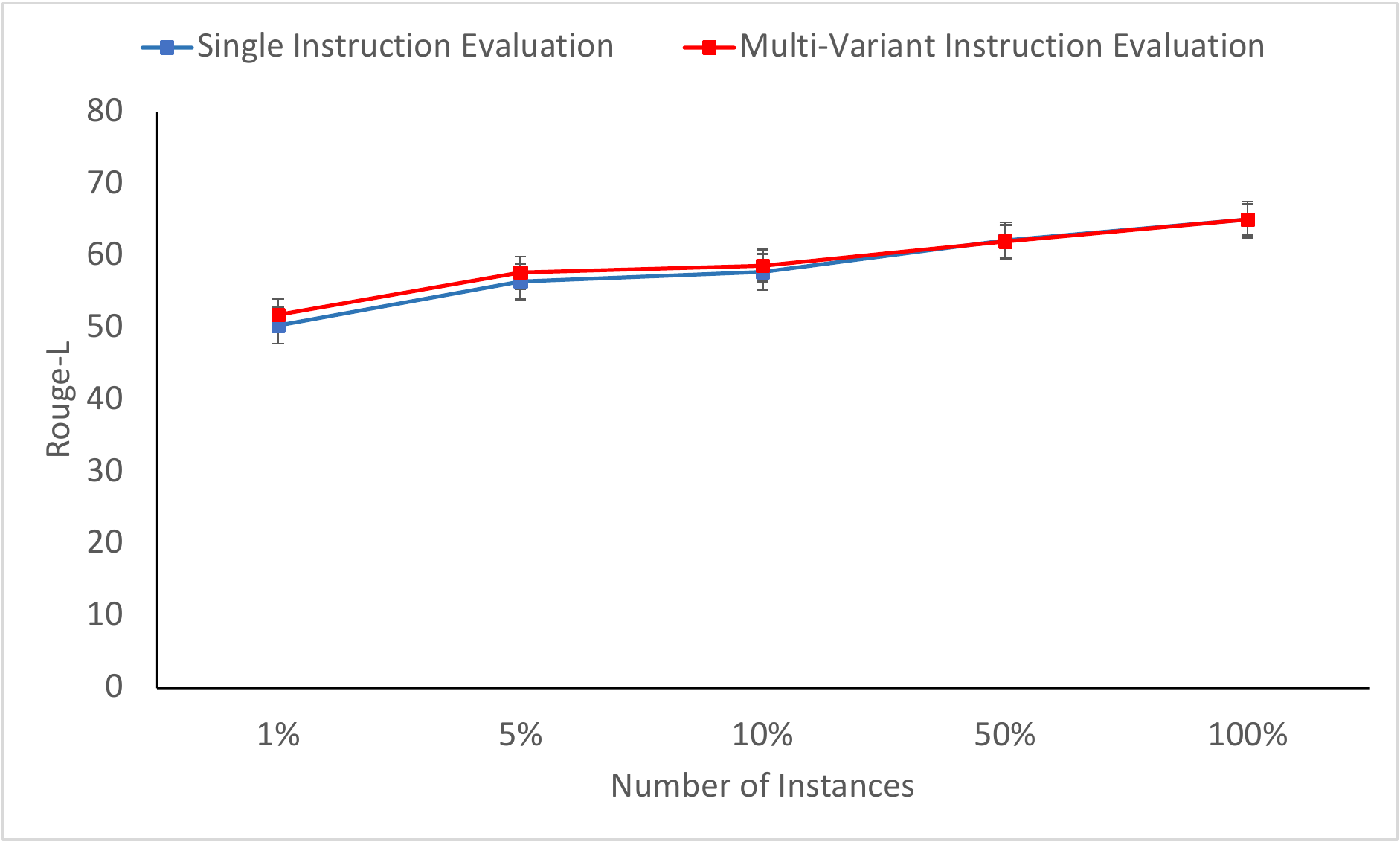} 
    \caption{Multi-task MVI learning} 
    \label{robustness_line_chart:b} 
    \vspace{4ex}
  \end{subfigure}
  \caption{Robustness comparison of SI \textit{vs.} MVI in multi-task setting - LM fine-tuned using MVI learning is more robust to variants as compared to SI learning.}
  \label{app: robustness_line_chart} 
\end{figure*}

\paragraph{Is single-instruction learning robust?} As Figure \ref{app: robustness_line_chart} illustrates, LM fine-tuned with single-instruction learning or original setting is not robust to instructions written in a different way; this includes transformation techniques like paraphrasing, adding spelling mistakes, grammatical mistakes etc. Our experiment results show that model trained using the proposed multi-variant instruction learning technique is able to perform reasonably well and is robust to variant instructions in both multi-task setting, as evident by lower performance difference between single instruction evaluation and multi-variant instruction evaluation setup.

\section{Contribution of Individual Variants}
\paragraph{Do each of the variant instructions contribute equally towards performance gain?} To analyse the contribution of each of the variant instructions, we study the performance gain by adding a single variant instruction at one time. We perform this analysis in TS setting (task\_010) and MT setting and summarize the results in Table \ref{tab:table_variant_contribution_single_task} and Table \ref{tab:table_variant_contribution_multi_task} respectively. We observe that all variants do not contribute equally, e.g. MVI\_All above are often smaller than individual MVIs. Identifying optimal variants, however, will be a scope for future work.

\begin{table*}[t]
\small
\centering
\renewcommand{\arraystretch}{1.2}
{
\resizebox{\linewidth}{!}{
\begin{tabular}{@{}ccccccccc@{}}
\toprule

\multirow{2}{*}{\# of Instances} &  \multicolumn{2}{c}{SI} & \multicolumn{2}{c}{Perturbation 1} & \multicolumn{2}{c}{Perturbation 2} & \multicolumn{2}{c}{Perturbation 3}\\ 
\cmidrule(lr){2-9} 
~& Original & Ours & Original & Ours & Original & Ours & Original & Ours\\ 
\bottomrule
 1\% & 7.34 & 34.53 & 7.73 & 39.76 & 7.23 & 33.27 & 3.37 & 35.32 \\
 5\% & 32.01 & 62.61 & 25.90 & 60.22 & 29.51 & 63.52 & 23.50 & 69.30 \\
 10\% & 46.93 & 63.61 & 46.36 & 61.70 & 44.74 & 63.86 & 43.28 & 72.46 \\
 50\% & 63.38 & 66.16 & 61.63 & 64.50 & 63.73 & 66.40 & 71.79 & 67.99 \\
 100\% & 64.99 & 67.15 & 63.12 & 67.38 & 65.05 & 66.02 & 72.70 & 68.24 \\
\bottomrule
\end{tabular}
}
}
\caption{Comparison of performance in multi-task setting across single-instruction and multi-variant instruction learning.}
\label{app:table_perturbation_results_multi_task}
\end{table*}

\begin{table*}[t]
    \setlength\tabcolsep{3.0pt}
    \setlength{\belowcaptionskip}{-10pt}
    \centering
    \footnotesize
\renewcommand{\arraystretch}{1.2}
{
\resizebox{0.9\linewidth}{!}{

\begin{tabular}{@{}ccccccccccc@{}}
\toprule
\multirow{3}{*}{\# of Instances} &  \multicolumn{4}{c}{BART-base} & \multicolumn{4}{c}{T5-base}\\
\cmidrule{2-9}
& \multicolumn{2}{c}{SI} & \multicolumn{2}{c}{MVI} & \multicolumn{2}{c}{SI} & \multicolumn{2}{c}{MVI} \\ 
\cmidrule{2-9}
~& Original & Ours & Original & Ours & Original & Ours & Original & Ours\\ 
\midrule
~& & & & 1\% tasks & & & & \\
\midrule
 1\% & 16.00 & 6.94 & 10.93 & 10.16 & 0.96 & 7.36 & 0.87 & 7.31\\
 5\% & 20.04 & 40.14 & 19.51 & 31.09 & 21.87 & 29.07 & 19.89 & 29.60\\
 10\% & 33.09 & 48.43 & 31.83 & 47.66 & 36.17 & 44.50 & 33.13 & 45.28\\
 50\% & 61.70 & 78.22 & 58.53 & 78.43 & 64.74 & 73.94 & 61.34 & 73.45 \\
 100\% & 68.66 & 84.22 & 64.39 & 84.87 & 72.35 & 83.37 & 68.9 & 84.2\\
\midrule
~& & & & 5\% tasks & & & & \\
\midrule
 1\% & 16.23 & 22.17 & 3.32 & 18.78 & 1.30 & 7.55 & 1.29 & 7.29\\
 5\% & 31.58 & 40.3 & 29.81 & 33.12 & 22.85 & 29.04 & 20.44 & 29.02\\
 10\% & 34.73 & 46.02 & 34.38 & 49.15 & 36.01 & 44.83 & 33.75 & 44.93\\
 50\% & 63.06 & 78.48 & 60.5 & 79.76 & 65.96 & 76.25 & 61.01 & 76.13\\
 100\% & 69.93 & 85.2 & 67.41 & 86.68 & 74.54 & 83.61 & 70.2 & 83.69\\
\midrule
~& & & & 10\% tasks & & & &\\
\midrule
 1\% & 2.98 & 22.16 & 2.46 & 19.98 & 3.12 & 7.89 & 2.56 & 7.66\\
 5\% & 29.27 & 30.06 & 28.03 & 30.9 & 24.49 & 29.29 & 23.41 & 29.25\\
 10\% & 39.95 & 46.38 & 36.3 & 50.4 & 36.76 & 45.22 & 36.23	& 44.81\\
 50\% & 63.58 & 79.13 & 59.98 & 79.81 & 66.07 & 73.49 & 62.56 & 73.54\\
 100\% & 70.82 & 86.66 & 69.11 & 87.86 & 71.97 & 81.16 & 70.34 & 81.08\\
\midrule
~& & & & 50\% tasks & & & &\\
\midrule
 1\% & 15.18 & 23.06 & 17.08 & 26.2 & 5.58 & 22.26 & 5.44 & 22.21 \\
 5\% & 32.88 & 44.5 & 33.88 & 44.64 & 33.56 & 40.37 & 30.57 & 38.25\\
 10\% & 43.33 & 51.2 & 42.5 & 54.62 & 45.42 & 44.02 & 39.01 & 44.36\\
 50\% & 68.18 & 80.8 & 66.42 & 81.29 & 66.62 & 80.97 & 63.89 & 80.93\\
 100\% & 71.35 & 84.52 & 68.85 & 84.65 & 72.72 & 82.82 & 69.94 & 82.02 \\
\midrule
~& & & & 100\% tasks & & & &\\
\midrule
 1\% & 17.04 & 22 & 19.2 & 24.95 & 20.69 & 22.55 & 9.02 & 20.66\\
 5\% & 35.4 & 42.68 & 36.42 & 45.06 & 35.18 & 38.30 & 30.92 & 39.51\\
 10\% & 46.4 & 60 & 45.33 & 59.3 & 44.70 & 53.80 & 44.47 & 54.15 \\
 50\% & 69.06 & 84.32 & 67.29 & 84.47 & 71.89 & 79.20 & 68.64 & 79.56\\
 100\% & 74.45 & 90.01 & 72.26 & 90.35 & 74.03 & 81.53 & 72.34 & 82.15\\
 \bottomrule
\end{tabular}
}
}
\caption{Comparison of performance in cross-task setting across single-instruction and multi-variant instruction learning. SI: Single-Instruction, MVI: Multi-Variant Instruction.}
\label{tab:table_cross_task}
\end{table*}
\begin{table*}[t]
\large
\centering
\resizebox{0.9\textwidth}{!}{
\begin{tabular}{@{}c|llllllllll@{}}
\toprule

\# of Instances & SI & MVI\_1 & MVI\_2 & MVI\_3 & MVI\_4 & MVI\_5 & MVI\_6 & MVI\_7 & MVI\_All \\ 
\bottomrule
 1\% & 0.00 & 17.46 & 0.92 & 0.20 & 0.44 & 6.92 & 5.7 & 6.79 & 0.00 \\
 5\% & 0.00 & 34.34 & 35.84 & 36.90 & 37.36 & 39.96 & 37.72 & 37.97 & 36.75\\
 10\% & 0.23 & 37.31 & 41.03 & 42.30 & 42.95 & 43.59 & 42.4 & 41.23 & 36.75\\
 50\% & 37.00 & 44.25 & 59.30 & 57.18 & 59.45 & 61.82 & 62.93 & 44.14 & 43.02\\
 100\% & 41.97 & 44.34 & 71.02 & 75.20 & 80.27 & 81.74 & 86.05 & 53.63 & 45.65\\
\bottomrule
\end{tabular}
}
\caption{Contribution of each variant instruction towards performance in task-specific setting for task010. SI: Single-Instruction, MVI\_k: Multi-Variant Instruction where k equals number of variant instructions used.}
\label{tab:table_variant_contribution_single_task}
\end{table*}
\begin{table*}[t]
\large
\centering
\resizebox{0.7\textwidth}{!}{
\begin{tabular}{@{}c|llllllllll@{}}
\toprule

\# of Instances & SI & MVI\_1 & MVI\_2 & MVI\_3 & MVI\_All \\ 
\bottomrule
 1\% & 15.84 & 37.03 & 40.93 & 64.08 & 50.4 \\
 5\% & 45.13 & 55.38 & 55.80 & 56.46 & 56.49\\
 10\% & 55.03 & 58.17 & 58.32 & 57.70 & 57.8\\
 50\% & 59.01 & 61.62 & 61.45 & 62.20 & 62.21\\
 100\% & 61.08 & 62.90 & 64.08 & 64.10 & 65.13\\
\bottomrule
\end{tabular}
}
\caption{Contribution of each variant instruction towards performance in multi-task setting. SI: Single-Instruction, MVI\_k: Multi-Variant Instruction where k equals number of variant instructions used.}
\label{tab:table_variant_contribution_multi_task}
\end{table*}

\end{document}